\documentclass{article}

\PassOptionsToPackage{numbers, compress}{natbib} 
\usepackage[final]{neurips_2026} 
\makeatletter
\def\@trackname{}
\makeatother
\usepackage[utf8]{inputenc} 
\usepackage[T1]{fontenc}    
\usepackage{hyperref}       
\usepackage{url}            
\usepackage{booktabs}       
\usepackage{amsfonts}       
\usepackage{nicefrac}       
\usepackage{microtype}      
\usepackage{xcolor}         

\usepackage{graphicx}
\usepackage{subcaption}
\usepackage{amsmath}
\usepackage{amssymb}
\usepackage{mathtools}
\usepackage{amsthm}
\usepackage{listings}
\definecolor{codebg}{HTML}{F5F8FC}
\definecolor{codeframe}{HTML}{D6E0EA}
\usepackage{enumitem}
\usepackage{multirow}
\usepackage{float}
\usepackage{tabularx}
\usepackage{makecell}
\usepackage{colortbl}
\usepackage[capitalize,noabbrev]{cleveref}
\usepackage[textsize=tiny]{todonotes}

\colorlet{hdrfill}{blue!12}
\colorlet{allfill}{green!10}
\colorlet{agentfill}{orange!6}

\theoremstyle{plain}

\theoremstyle{definition}

\theoremstyle{remark}

\title{PDEAgent-Bench: A Multi-Metric, Multi-Library Benchmark for PDE Solver Generation}

\author{%
\renewcommand{\arraystretch}{1.15}
\begin{tabular}{c}
Zhen Hang$^{1}$\thanks{Equal contribution.}\quad
Yushan Yashengjiang$^{1}$\footnotemark[1]\quad
Junhui Li$^{2}$\quad
Huanshuo Dong$^{1}$\quad
Yang Wei$^{3}$\\
Zhezheng Hao$^{4}$\quad
Jiangtao Ma$^{5}$\quad
Songlin Bai$^{12}$\quad
Haozhong Kai$^{6}$\quad
Xihang Yue$^{4}$\\
Gangzong Si$^{1}$\quad
Dongming Jiang$^{7}$\quad
Chao Yao$^{8}$\quad
Zhanhua Hu$^{9}$\quad
Jiangqing Zhang$^{3}$\\
Pengwei Liu$^{4}$\quad
Yaomin Shen$^{4}$\quad
Xingyu Ren$^{4}$\quad
Lei Liu$^{1}$\quad
Zikang Xu$^{1}$\quad
Han Li$^{10}$\\
Qingsong Yao$^{12}$\quad
Hande Dong$^{1}$\quad
Hong Wang$^{1}$\thanks{Corresponding author. Contact: \texttt{wanghong1700@mail.ustc.edu.cn}.}\\
\\[-0.3ex]
$^{1}$University of Science and Technology of China \\
$^{2}$Beijing University of Posts and Telecommunications \\
$^{3}$Shanghai Jiao Tong University \quad 
$^{4}$Zhejiang University \quad
$^{5}$National University of Singapore \\
$^{6}$Tsinghua University \quad
$^{7}$University of Texas at Dallas \quad
$^{8}$Arizona State University \\
$^{9}$Rice University \quad
$^{10}$Technical University of Munich \\
$^{11}$Stanford University \quad
$^{12}$Alibaba Group
\end{tabular}%
}

\begin{document}

\maketitle
\begin{abstract}
PDE-to-solver code generation aims to automatically synthesize executable numerical solvers from partial differential equation (PDE) specifications.
This task requires not only understanding the mathematical structure of PDEs, but also selecting appropriate discretization schemes and solver configurations, and correctly implementing the resulting formulations in finite-element method (FEM) libraries. Existing code generation benchmarks mainly evaluate syntactic correctness, or success on predefined test cases. To our knowledge, there is currently no publicly available benchmark specifically for PDE-to-solver code generation, and general-purpose code benchmarks do not fully capture the unique challenges of numerical PDE solution, such as ensuring solver accuracy, efficiency, and compatibility with professional FEM libraries.

We introduce PDEAgent-Bench, to the best of our knowledge, the first multi-metric, multi-library benchmark for PDE-to-solver code generation. PDEAgent-Bench contains 645 instances across 6 mathematical categories and 11 PDE families, with common FEM libraries for DOLFINx, Firedrake, and deal.II. 
Each instance provides an agent-facing problem specification, a reference solution on a prescribed evaluation grid, and case-specific accuracy and runtime targets. 
PDEAgent-Bench adopts a staged evaluation framework in which generated solvers must sequentially pass executability, numerical accuracy, and computational efficiency checks. 
Experiments with representative LLMs and code agents show that models can often produce runnable code, but their pass rate drops substantially once accuracy and efficiency requirements are enforced. These results indicate that current agents remain limited in producing numerically reliable and efficient PDE solvers, and that PDEAgent-Bench provides a reproducible testbed grounded in the practical requirements of numerical PDE solving. Dataset, scripts, codes, and the project homepage are publicly released at
\href{https://github.com/YusanX/pde-agent-bench}{\texttt{https://github.com/YusanX/pde-agent-bench}}
and
\href{https://zeroeclipse00.github.io/pde-agent-bench-github-pages/#home}{\texttt{https://zeroeclipse00.github.io/pde-agent-bench-github-pages/}}.

\end{abstract}

\section{Introduction}
Numerical solvers for partial differential equations (PDEs) are central to scientific computing and industrial simulation, with applications in fluid dynamics, structural mechanics, heat transfer, electromagnetics, and materials modeling.
Constructing reliable PDE solvers, however, is not merely a coding task: it requires translating mathematical models and boundary conditions into stable discretizations, weak forms or time-stepping schemes, and properly configured linear or nonlinear solvers.
Mistakes in these choices can produce programs that run but impose incorrect physical constraints, fail to converge, yield inaccurate solutions, or exceed practical resource budgets.


\begin{figure}[t]
    \centering
    \includegraphics[width=\linewidth]{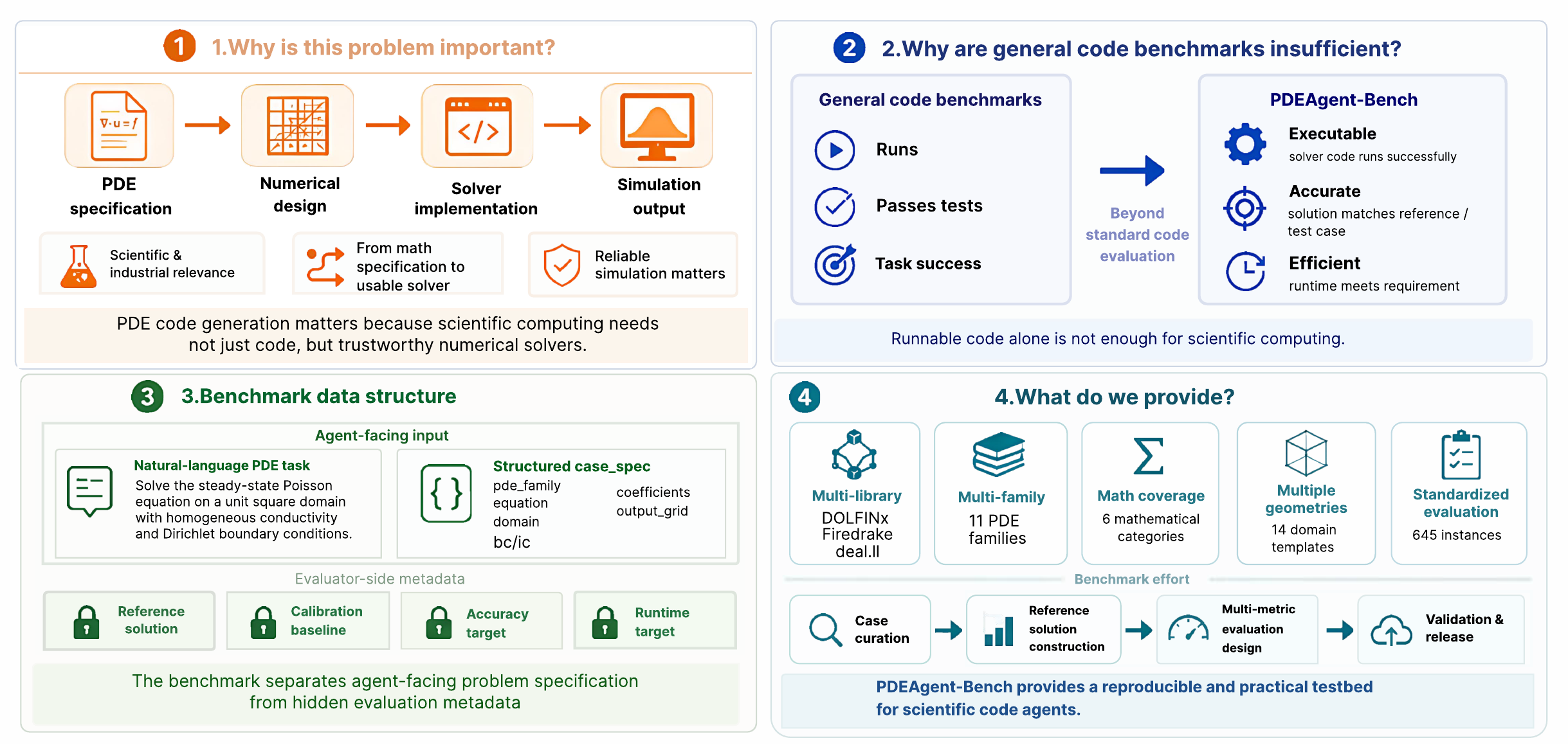}
    \caption{Overview of PDEAgent-Bench: motivation, limitations of general code benchmarks, benchmark data structure, and evaluation protocol.}
    \vspace{-5mm}

    \label{fig:intro}
\end{figure}

Benchmarks for code generation agents already span a wide range of settings, from function-level programming problems to repository-level task solving. For example,
HumanEval~\cite{chen2021evaluating}, MBPP~\cite{austin2021programsynthesislargelanguage}, APPS~\cite{hendrycks2021measuring}, and CodeContests~\cite{li2022competition} primarily evaluate code generation on programming tasks, while SWE-bench~\cite{jimenez2024swebench} and Terminal-Bench~\cite{merrill2026terminalbench} further assess repository modification and task execution in interactive terminal environments.
These benchmarks typically rely on unit tests, task-completion signals, or patch validation, making them well suited for measuring discrete functional correctness.
However, PDE solver synthesis is not evaluated by whether isolated tests pass, but by whether the generated program actually solves a specified PDE, which requires assessing numerical solution quality and computational cost.

A benchmark for PDE-to-solver code generation therefore requires a stricter numerical evaluation protocol.
Each instance must provide not only agent-facing PDE specifications, but also reliable reference solutions, prescribed evaluation grids, and consistent procedures for measuring error and runtime, enabling reproducible comparison across models and implementations.
The evaluation must also distinguish execution failures, excessive numerical error, and insufficient computational efficiency, since a program may run successfully without correctly solving the target PDE.
To reflect realistic scientific-computing practice, the benchmark should further cover solver implementations across professional FEM software stacks such as DOLFINx~\cite{BarattaEtal2023dolfinx}, Firedrake~\cite{Firedrake}, and deal.II~\cite{Bangerth2007DealII}.

To address this gap, we introduce PDEAgent-Bench, the first multi-metric, multi-library benchmark for PDE-to-solver code generation by LLMs and code agents.
PDEAgent-Bench contains 645 instances spanning 6 mathematical categories and 11 representative PDE families, with library-specific tracks for DOLFINx, Firedrake, and deal.II.
Each instance includes an agent-facing PDE specification, a prescribed evaluation grid, a reference solution, and case-specific accuracy and runtime targets to support reproducible numerical verification.
By evaluating generated solvers across multiple FEM software stacks, PDEAgent-Bench assesses model capabilities across different scientific-computing libraries and implementation paradigms, rather than within a single API environment.

PDEAgent-Bench adopts a staged evaluation framework centered on case-level pass rate.
Generated solvers are executed in a controlled environment and must sequentially pass executability, numerical accuracy, and computational efficiency checks.
Beyond the final pass rate, we report stage-level and auxiliary metrics to diagnose failure sources and analyze solution quality.
Experiments with state-of-the-art LLMs and code agents show that current systems can generate runnable code for a subset of PDE instances, but their performance drops substantially once numerical and resource constraints are enforced, with clear library-dependent failure modes.

Our contributions are summarized as follows:
\begin{itemize}
    \item We introduce PDEAgent-Bench, to the best of our knowledge, the first benchmark for evaluating whether LLMs and code agents can synthesize executable numerical solvers from PDE specifications, extending code-agent evaluation beyond discrete functional correctness to include numerical accuracy and computational efficiency.

    \item We construct a collection of 645 PDE instances spanning 6 mathematical categories and 11 representative PDE families, with specifications, evaluation grids, reference solutions, and case-specific targets for reproducible verification.

    \item We provide library-specific tracks for DOLFINx, Firedrake, and deal.II, covering Python and C++ FEM ecosystems and evaluating solver generation across scientific-computing software stacks.

    \item We design a staged evaluation protocol centered on case-level pass rate, with stage-level and auxiliary metrics for diagnosing executability, numerical accuracy, computational efficiency, and solution quality.
\end{itemize}



\begin{figure}[t]
    \centering
    \includegraphics[width=\linewidth]{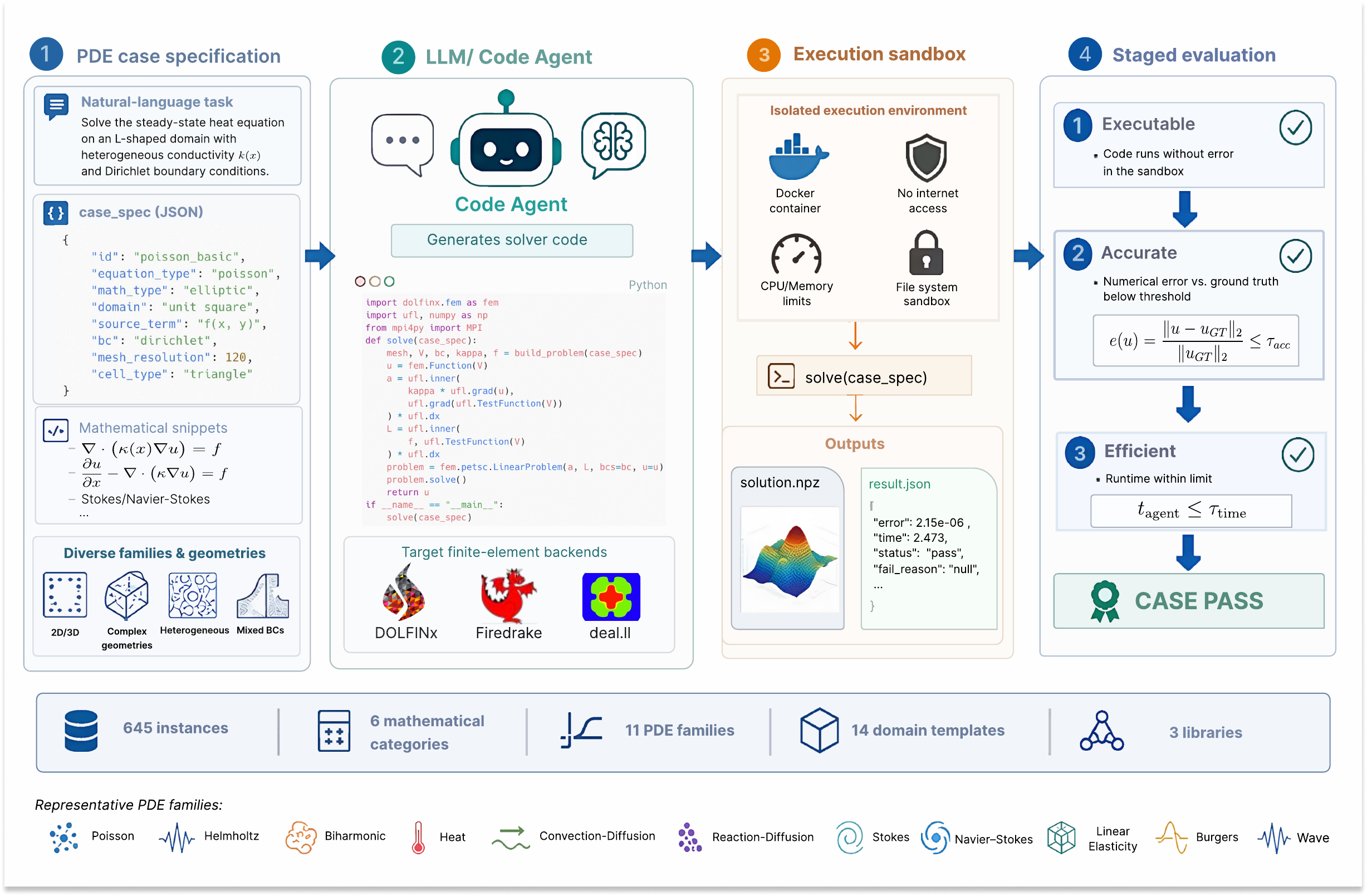} 
    \caption{PDEAgent-Bench evaluation pipeline from PDE case specification to sandboxed execution and staged assessment.}
    \label{fig:method}
\end{figure}

\section{Benchmark Overview}
\label{sec:task}

\subsection{Task Scope and Evaluation Setting}

PDEAgent-Bench evaluates PDE-to-solver code generation: given a PDE or physics specification (input), a system must generate an executable numerical solver (output) that reaches target accuracy within practical resource budgets (success criteria). 
Our goal is not to propose a new solver-generation method, but to evaluate the capabilities, limitations, and failure modes of existing LLMs and code agents on this task.

Each PDE problem defines a single evaluation instance. 
An instance abstracts the task faced by the code agent, separating agent-visible information from hidden evaluator-only details. 
The agent interacts with a problem description and structured case specification, while the evaluator retains reference solutions and other metadata to measure performance. 
Evaluation proceeds via a common interface with staged checks for executability, numerical accuracy, and runtime efficiency.

PDEAgent-Bench contains 645 instances across 6 mathematical categories and 11 representative PDE families. 
It provides three FEM-library tracks: DOLFINx as the full-coverage primary track, and Firedrake and deal.II as cross-library tracks. 
These tracks evaluate whether code agents can transfer solver-synthesis capability across different FEM ecosystems, including both Python- and C++-based toolchains.

\subsection{Task Specification and Evaluation Interface}

Each instance contains a structured problem specification and a corresponding natural-language description.
The structured specification defines the PDE family, computational domain, coefficient functions, source terms, boundary conditions, initial conditions, and temporal parameters when applicable.
The natural-language description is generated from this specification and summarizes the PDE type, physical context, and computational requirements, including the evaluation grid and optional numerical hints.
Together, they form the agent-facing task input, ensuring that different systems generate solvers from the same PDE information.

The submitted solver receives the case specification through a unified interface and returns a numerical solution on the evaluation grid.
Agents may choose their own internal mesh, discretization method, time-stepping scheme, and solver settings.
Evaluation only requires that the returned solution have the correct output format and satisfy the subsequent accuracy and runtime checks.
This interface allows diverse numerical implementation strategies while ensuring that all submissions are compared through a common output format.

Evaluation is performed in an isolated environment with fixed resource limits.
The evaluator invokes the generated solver, records elapsed runtime, and validates the format and dimensions of the returned solution.
Reference solutions, calibration baselines, and hidden numerical settings used to generate them are not exposed to the agent.
When reference solutions are constructed from analytic expressions, those expressions are used only internally to generate reference data and consistent PDE terms, and are not provided as target solution fields.
This information separation ensures that all submissions are evaluated under the same conditions.

\subsection{Evaluation Framework and Metrics}
\label{sec:scoring_leaderboard}

PDEAgent-Bench distinguishes the reference solution used for scoring from the calibration baseline used for threshold setting.
The reference solution, denoted by $u_{\mathrm{GT}}$, is used to compute the error of a submitted solution.
When the reference solution can be constructed from an analytic expression, we sample $u_{\mathrm{GT}}$ directly on the evaluation grid; otherwise, we generate it using high-accuracy numerical settings, such as finer meshes, higher-order discretizations, and stricter solver tolerances.
The calibration baseline, denoted by $u_{\mathrm{base}}$, is a separately generated numerical solution used only to set case-specific accuracy and runtime thresholds, and is not included in model comparisons.

For any submitted solution $u$, we compute the relative $L^2$ error on the evaluation grid and use it as the primary accuracy metric; 
if the reference solution is identically zero, we instead use the absolute $L^2$ error to avoid division by zero.
\begin{equation}
e(u)=
\frac{\|u-u_{\mathrm{GT}}\|_2}
     {\|u_{\mathrm{GT}}\|_2}.
\label{eq:relative_l2}
\end{equation}
Let $e_{\mathrm{base}}=e(u_{\mathrm{base}})$ denote the calibration error and let $t_{\mathrm{base}}$ denote the calibration-baseline runtime.
Because numerical difficulty and computational cost vary substantially across PDE instances, PDEAgent-Bench adopts calibration-based, case-specific thresholds. 
To reduce variability from short runtimes or environmental noise, each solver is executed multiple times and the runtime is averaged before comparison.
\begin{equation}
\tau_{\mathrm{acc}} = \max\!\left(\alpha_{\mathrm{acc}}\,e_{\mathrm{base}},\;\tau_{\min}\right),
\qquad
\tau_{\mathrm{time}} = \alpha_{\mathrm{time}} t_{\mathrm{base}}.
\label{eq:thresholds}
\end{equation}
The accuracy threshold applies a fixed tolerance multiplier to the calibration error, allowing reasonable implementation differences such as mesh resolution, discretization order, or solver settings, while a minimum floor avoids overly strict thresholds when the calibration error is near zero.
The runtime threshold similarly permits a bounded overhead relative to the calibration-baseline runtime.
By default, we set $\alpha_{\mathrm{acc}}=10$, $\alpha_{\mathrm{time}}=3$, and $\tau_{\min}=10^{-6}$; these values are fixed across all models and instances, with threshold distributions and sensitivity analyses reported in Appendix~\ref{app:calibration_distribution} and Appendix~\ref{app:threshold_sensitivity} .

Each library-instance is evaluated in stages.
A submission must first execute successfully and return a result with valid format and dimensions; it must then satisfy the accuracy condition $e(u_{\mathrm{agent}})\leq \tau_{\mathrm{acc}}$ and the runtime condition $t_{\mathrm{agent}}\leq \tau_{\mathrm{time}}$.
A library-instance is counted as passed only when executability, numerical accuracy, and runtime all meet the required criteria.
We summarize model performance using the case-level pass rate:
\begin{equation}
\mathrm{pass\_rate}(m)
=
\frac{1}{|\mathcal{D}|}
\sum_{i\in\mathcal{D}}
\mathrm{case\_pass}_{i}(m),
\label{eq:pass_rate}
\end{equation}
where $\mathcal{D}$ denotes the set of library-instances in the corresponding FEM-library track or evaluation split, and $\mathrm{case\_pass}_{i}(m)$ indicates whether model $m$ passes the $i$-th library-instance.

In addition to the final pass rate, PDEAgent-Bench reports execution pass rate, accuracy pass rate, and runtime pass rate to identify failures at different evaluation stages.
We also record auxiliary diagnostic metrics for analyzing solution quality, discretization choices, solver behavior, stability-related properties, and efficiency patterns.
These auxiliary metrics do not affect pass/fail decisions; their full definitions and results are provided in the appendix.






\section{Dataset Construction}
\label{sec:dataset}
PDEAgent-Bench consists of 645 PDE instances covering diverse PDE families, mathematical problem types, and spatial geometries. 
The dataset is built through a controlled generation-and-validation pipeline: we first define family-level design axes to cover different numerical regimes, and then use structured templates with LLM-assisted generation to instantiate candidate \texttt{case\_spec} records describing the core PDE setting, computational domain, boundary conditions, and evaluation grid. 
Each candidate is paired with a reference solution, and a baseline solver is used to calibrate case-specific accuracy and runtime thresholds. 
Only candidates that pass schema checks and trial execution are included in the final dataset. 
The released records separate agent-visible task fields from evaluator-only reference and calibration metadata. 
The internal structure of each instance is illustrated in Figure~\ref{fig:intro}, with full construction details and the procedure for adding new FEM-library tracks provided in Appendix~\ref{app:pipeline}.

\subsection{PDE Families and Domain Coverage}
\label{subsec:taxonomy}
Each instance is labeled by its PDE family and a coarse-grained mathematical type. The benchmark spans 11 representative PDE families covering common numerical-simulation settings—including elliptic, parabolic, hyperbolic, incompressible-flow, reaction-diffusion, and elasticity problems—and captures six mathematical categories reflecting dominant operator structures. Instances are designed to challenge solvers with numerical difficulties such as strong coefficient contrast, nonlinear terms, coupled variables, and stability-sensitive discretizations, testing capabilities beyond basic elliptic problems.

The dataset also systematically covers diverse spatial geometries.
It contains 14 domain templates, ranging from simple domains such as the unit square and unit cube to more complex geometries with nonconvex structures, curved boundaries, multiply connected regions, re-entrant corners, periodic boundaries, and irregular boundary tagging.

For each domain, the output grid is defined on its bounding box; grid points outside the physical domain are marked as NaN and excluded from error computation.
Submissions must therefore handle out-of-domain sampling points, rather than relying only on simple rectangular-grid assumptions.

\subsection{Instance Format}
\label{subsec:schema}

Each instance is stored as a JSONL record.
The central agent-facing field is \texttt{case\_spec}, which specifies the PDE, computational domain, boundary and initial conditions, coefficient functions, temporal parameters, and evaluation grid.
The record also contains lightweight \texttt{tags} that describe structural properties and expected numerical difficulty for dataset analysis.
Reference solutions, calibration baselines and other construction-time information are stored as evaluator-side metadata and are not included in the agent-facing task input.

During evaluation, only \texttt{case\_spec} is included in the agent prompt and passed to the generated solver through the unified interface.
Evaluator-side metadata is used only for reference-solution construction, calibration-baseline generation, threshold setting, and reproducibility.
For instances whose reference solutions are constructed from analytic expressions, those expressions are used internally to generate consistent source terms, boundary conditions, or initial conditions; the agent-facing specification contains only the resulting PDE-defining information and does not expose the target solution field.
A representative JSONL record is provided in Appendix~\ref{app:schema}.

\subsection{Extensibility and Validation}
\label{subsec:pipeline}

PDEAgent-Bench is designed to support controlled extensions under the same evaluation protocol.
To add a new instance, an author provides a structured PDE specification together with the corresponding domain, coefficients, boundary and initial conditions, temporal parameters, and output grid.
The construction pipeline then builds or verifies the reference solution, runs the calibration baseline to obtain the calibration error and runtime, derives the case-specific thresholds, and exports the agent-facing specification, natural-language task description, library metadata, and evaluator files.

Before inclusion, each candidate instance must pass automated validation.
Static checks verify schema completeness, identifier uniqueness, expression parsability, consistency between labels and specifications, and validity of reference and calibration settings.
Trial execution verifies that reference generation and calibration runs are stable, that the generated evaluation files can be invoked by the evaluator, and that output-format checking and error computation behave correctly.
We provide instance and FEM-library-track templates in the appendix and open-source release to support future extensions.

\section{Experiments}
\label{sec:experiments}
\paragraph{Experiment setup.}
All experiments use the staged evaluation framework defined in Section~\ref{sec:scoring_leaderboard}.
Generated solvers are executed on a single-node CPU server (AMD EPYC 7K62, 32 visible cores, 62~GB RAM) running Ubuntu 22.04, with a per-case timeout of 300~s.
The software stack uses Python 3.11, GCC 11.4, MPICH, PETSc 3.24, DOLFINx 0.10.0, Firedrake (2025), and deal.II 9.7.
Full hardware, container-image, dependency, and resource-limit details are provided in Appendix~\ref{app:reproducibility}.

\paragraph{Baseline models.}
We evaluate two classes of systems: base LLMs that directly generate complete solvers, and Code/PDE agents that use code-generation, editing, or execution workflows. The base LLMs include GPT-5.4, Opus 4.7, Gemini 3.1 Pro, Qwen3.6-Plus, and DeepSeek V3.2; the agent systems include CodePDE~\cite{li2025codepde}, OpenHands~\cite{wang2024openhands}, and Mini-SWE~\cite{yang2024sweagent}. To separate the effect of agentic scaffolding from backbone model capability, all three agent systems use GPT-5.4 as the underlying model. All model calls use sampling temperature $0$. Additional prompt, hardware, dependency, and resource-limit details are provided in Appendix~\ref{app:reproducibility}.


\subsection{Single-shot Solver Synthesis}
\label{sec:singleshot}

We first evaluate the \emph{single-shot} setting as the basic baseline for PDE-to-solver generation. In this setting, each model--FEM library--instance pair produces exactly one solver submission and receives no execution feedback before submission, such as compilation errors, runtime errors, numerical errors, or runtime measurements. This setting therefore measures the ability of a system to complete the numerical formulation, FEM library implementation, and solver configuration in a single attempt. Base LLMs directly generate a complete solver from the structured case description, while Code/PDE agents may use their own code-generation or editing workflows but cannot perform feedback-driven repair based on execution results. The single-shot setting also serves as the main reference point for the execution-feedback and template-guided experiments in later subsections.

\paragraph{Overall ranking and capability gap.}
Table~\ref{tab:eqtype_backend_pass_rates} reports single-shot pass rates across all models and FEM-library tracks. On the primary DOLFINx track, Gemini 3.1 Pro achieves the highest overall pass rate at $\mathbf{54.1\%}$, followed by Opus 4.7 ($47.8\%$), GPT-5.4 ($46.0\%$), CodePDE ($44.7\%$), and OpenHands ($43.6\%$). The narrow spread among these top-performing systems suggests that, in the no-feedback single-shot setting, agent-style scaffolding does not yield a consistent advantage over direct generation by strong base LLMs. A substantial gap opens below this cluster: Qwen3.6-Plus reaches $23.4\%$, while DeepSeek V3.2 ($4.2\%$) and Mini-SWE ($3.7\%$) fail on nearly all instances. These results underscore that PDE solver synthesis requires more than code executability: models must also choose suitable numerical formulations, FEM-library APIs, discretizations, and solver configurations.

\begin{table*}[!htbp]
\centering
\scriptsize
\setlength{\tabcolsep}{4pt}
\renewcommand{\arraystretch}{1.25}
\newcommand{\rh}[1]{\rotatebox{55}{\textbf{#1}}}
\resizebox{\textwidth}{!}{%
\begin{tabular}{@{} ll l *{11}{c} >{\columncolor{allfill}}c @{}}
\toprule
\rowcolor{hdrfill}
\textbf{FEM library} & \textbf{Type} & \textbf{Model} &
\rh{Biharmonic} & \rh{Conv.-Diff.} & \rh{Heat} & \rh{Helmholtz} &
\rh{Lin.\ Elast.} & \rh{Stokes} & \rh{Poisson} &
\rh{React.-Diff.} & \rh{N.-Stokes} & \rh{Burgers} & \rh{Wave} &
\textbf{All} \\
\midrule

\multirow{9}{*}{\textbf{DOLFINx}}
& \multirow{6}{*}{\makecell[l]{Base\\LLM}}
& GPT-5.4        & 47.4 & 46.4 & 48.0 & 46.8 & 47.6 & \textbf{34.4} & 47.3 & 46.9 & 50.0 & 46.5 & 47.6 & 46.0 \\
& & Opus 4.7       & 50.9 & \textbf{58.3} & 26.0 & 62.9 & \textbf{66.7} & 24.6 & \textbf{52.7} & 51.6 & \textbf{57.1} & 27.9 & 28.6 & 47.8 \\
& & Gemini 3.1 Pro & \textbf{77.2} & 52.4 & \textbf{50.0} & \textbf{71.0} & \textbf{66.7} & 21.3 & 51.6 & \textbf{64.1} & 21.4 & \textbf{53.5} & \textbf{47.6} & \textbf{54.1} \\
& & Qwen3.6-Plus      & 24.6 &  29.8 &  34.0 &  40.3 &  28.6 &  6.6 & 16.5 &  17.2 &  17.9 & 18.6 & 21.4 &  23.4 \\
& & DeepSeek V3.2  &  8.8 &  2.4 & 10.0 &  0.0 &  3.2 &  1.6 & 13.2 &  0.0 &  0.0 &  0.0 &  0.0 &  4.2 \\
\cmidrule(l){2-15}
& \multirow{3}{*}{\makecell[l]{Code/PDE\\Agent}}
& CodePDE        & 38.6 & 45.2 & 50.0 & 69.4 &61.9 &19.7 &39.6 &56.2 &32.1 &41.9 &23.8 &44.7 \\
& & OpenHands     & 31.6 & 50.0 & 64.0 & 59.7 & 38.1 & 16.4 & 47.3 & 53.1 & 21.4 & 51.2 & 31.0 & 43.6 \\
& & Mini-SWE      & 5.3 & 7.1 & 2.0 & 0.0 & 6.3 & 1.6 & 3.3 & 3.1 & 3.6 & 4.7 & 2.4 & 3.7 \\

\bottomrule
\end{tabular}%
}
\caption{Single-shot case-level pass rates (\%) by PDE family on the primary DOLFINx track.}
\label{tab:eqtype_backend_pass_rates}
\end{table*}

\begin{figure}[!htbp]
\centering
\begin{subfigure}[b]{0.48\textwidth}
  \centering
  \includegraphics[width=\linewidth]{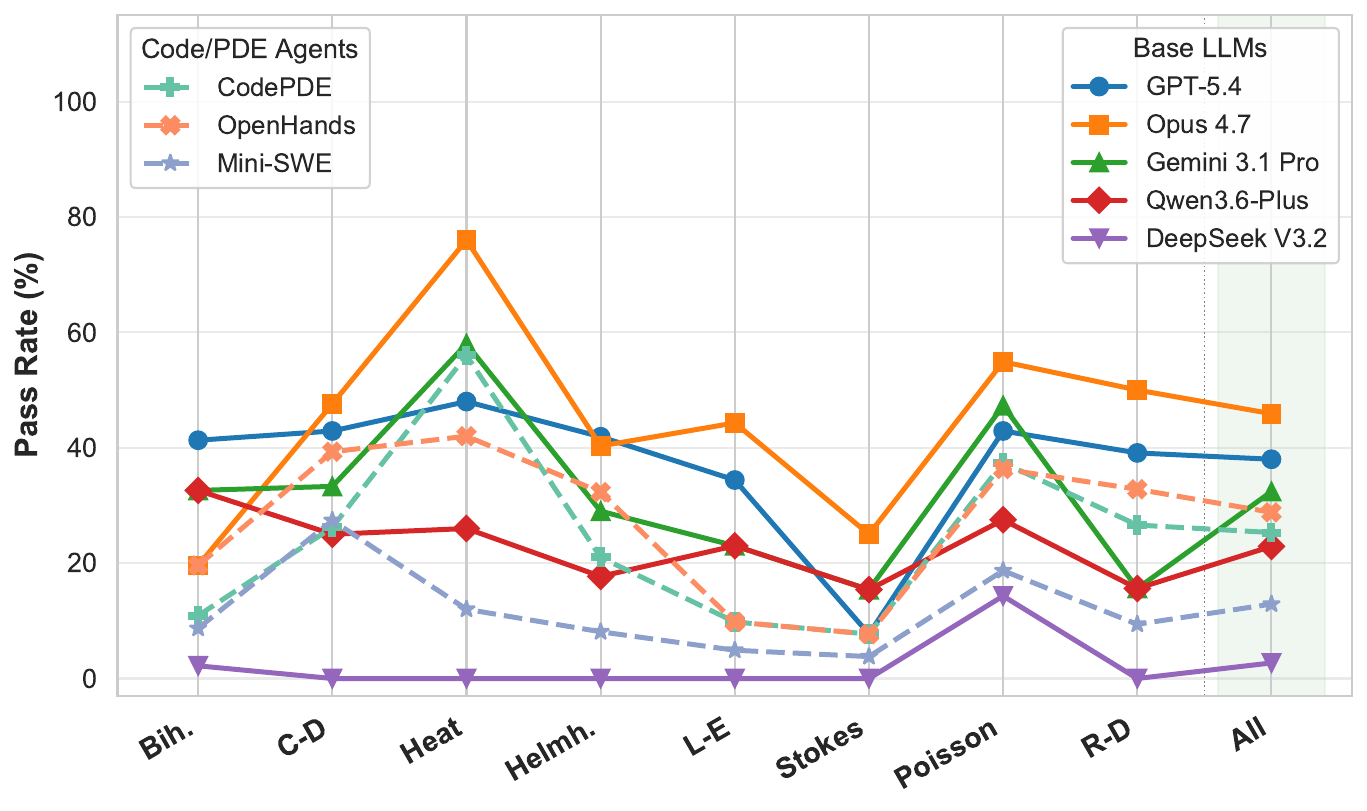}
  \caption{Firedrake}
  \label{fig:eqtype_firedrake}
\end{subfigure}
\hfill
\begin{subfigure}[b]{0.48\textwidth}
  \centering
  \includegraphics[width=\linewidth]{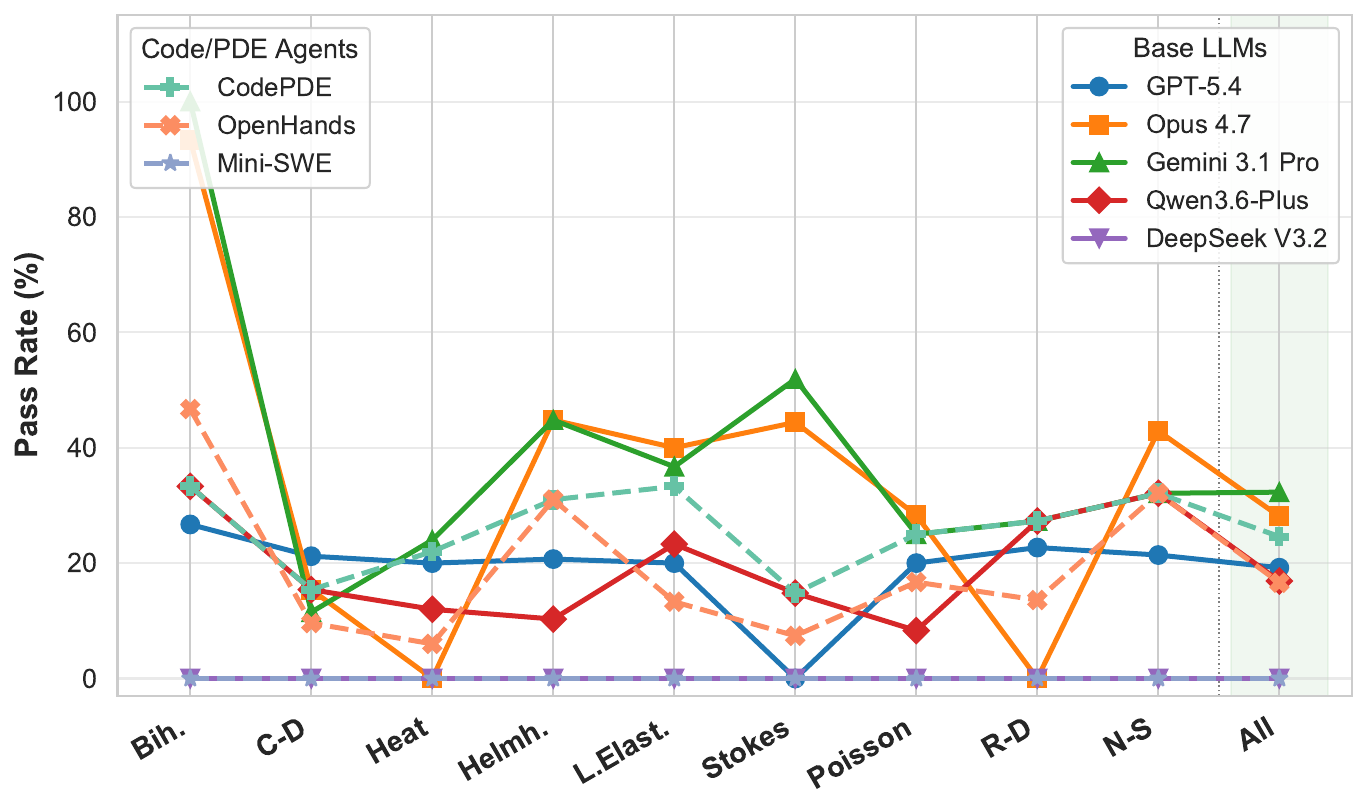}
  \caption{deal.II}
  \label{fig:eqtype_dealii}
\end{subfigure}
\caption{Single-shot pass rates (\%) by PDE family on Firedrake (left) and deal.II (right).
Solid lines denote base LLMs, and dashed lines denote Code/PDE agents.
Unavailable family--library pairs are omitted.
Full numbers are reported in Table~\ref{tab:eqtype_firedrake_dealii} in the Appendix.}
\label{fig:backend_eqtype}
\end{figure}

\paragraph{Library transfer and PDE-family effects.}
Figure~\ref{fig:backend_eqtype} and Table~\ref{tab:eqtype_backend_pass_rates} show that single-shot performance depends strongly on both the target FEM library and the PDE family. Pass rates drop substantially on the Firedrake and deal.II tracks relative to DOLFINx. Under the current FEM-library track composition, deal.II yields the lowest overall pass rates: even the strongest model on this track, Gemini 3.1 Pro, reaches only $32.3\%$. This indicates that cross-library transfer remains a major bottleneck, especially when moving from Python-based FEM libraries to the C++-based deal.II library. Family-level results reveal similarly persistent structure. Stokes is among the hardest families on DOLFINx: most systems remain below $25\%$, with only GPT-5.4 reaching a non-trivial $34.4\%$. This suggests that saddle-point structure, pressure null-space handling, and inf-sup compatibility remain difficult for current systems. Other families expose model-specific strengths: Gemini 3.1 Pro reaches $77.2\%$ on Biharmonic, while Opus 4.7 leads on Helmholtz and Linear Elasticity, indicating that capability is not uniform across PDE types.

\subsection{Failure-Stage Analysis}
\label{sec:failure_analysis}

\begin{figure}[!htbp]
\centering
\begin{subfigure}[b]{0.32\textwidth}
  \centering
  \includegraphics[width=\linewidth]{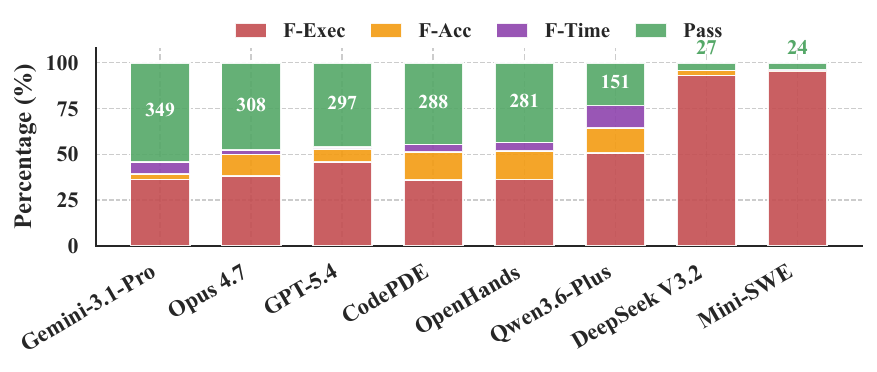}
  \caption{DOLFINx}
  \label{fig:failure_DOLFINx}
\end{subfigure}
\hfill
\begin{subfigure}[b]{0.32\textwidth}
  \centering
  \includegraphics[width=\linewidth]{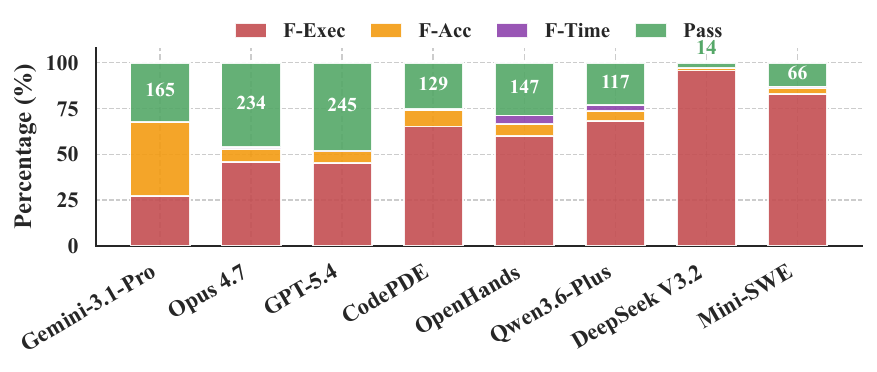}
  \caption{Firedrake}
  \label{fig:failure_firedrake}
\end{subfigure}
\hfill
\begin{subfigure}[b]{0.32\textwidth}
  \centering
  \includegraphics[width=\linewidth]{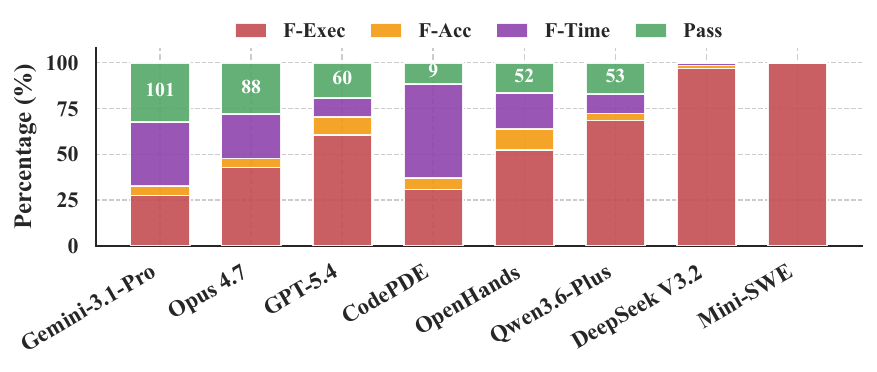}
  \caption{deal.II}
  \label{fig:failure_dealii}
\end{subfigure}
\caption{Failure-stage breakdown by model and FEM library. Each bar is partitioned into execution failure (F-Exec), accuracy failure (F-Acc), timeout failure (F-Time), and Pass.}
\label{fig:failure_stages}
\end{figure}

Figure~\ref{fig:failure_stages} decomposes the single-shot outcomes into four mutually exclusive categories: Pass, execution failure (F-Exec), accuracy failure (F-Acc), and timeout failure (F-Time). F-Exec denotes cases where the generated program fails to run successfully or returns an invalid output format; F-Acc denotes executable cases whose numerical error exceeds the accuracy threshold; and F-Time denotes cases that pass execution and accuracy checks but exceed the runtime threshold. Compared with the final pass rate, this breakdown more directly shows where models fail in the staged evaluation.

Overall, F-Exec is the dominant failure source for most models and library tracks. DeepSeek V3.2 and Mini-SWE fail almost entirely at the execution stage, indicating that they still struggle to generate runnable code for complex FEM libraries. Stronger systems, including Gemini 3.1 Pro, Opus 4.7, GPT-5.4, CodePDE, and OpenHands, show more distributed failure profiles: besides execution failures, F-Acc and F-Time also account for a visible fraction of outcomes. This suggests that once models achieve basic executability, numerical accuracy and computational efficiency become the next bottlenecks.

The failure profile also varies across FEM libraries. 
On deal.II, Gemini 3.1 Pro and Opus 4.7 show elevated F-Time rates of $35.1\%$ and $24.3\%$, respectively, indicating that solver configuration and resource efficiency remain challenging on the C++ library track. 
On Firedrake, Gemini 3.1 Pro has a high F-Acc rate of $40.2\%$, suggesting that many failures arise from numerical errors, potentially related to variational forms, boundary-condition handling, or output interpolation. 
Overall, the stage-level breakdown shows that weaker models are mainly limited by executability, whereas stronger models increasingly fail on accuracy and efficiency. 
Appendix~\ref{app:walkthroughs} provides representative case-level walkthroughs, including the agent-facing specification, hidden reference and calibration metadata, submitted-solver summary, numerical error, runtime, and final staged verdict.

\vspace{-0.4cm}

\subsection{Iterative Improvement with Execution Feedback}
\label{sec:iterative_feedback}

To assess whether execution feedback helps models repair generated solvers, we evaluate a three-attempt iterative protocol using GPT-5.4 as a representative strong model. For each library--instance pair, the model can submit up to three solvers. After each attempt, it receives execution feedback, such as standard error messages or structured stage-level outcomes, but never the reference solution. This setting tests which failures can be repaired from execution feedback and which still require correct numerical formulation and solver design.

\begin{figure}[!htbp]
\centering
\begin{subfigure}[b]{0.38\textwidth}
  \centering
  \includegraphics[width=\linewidth]{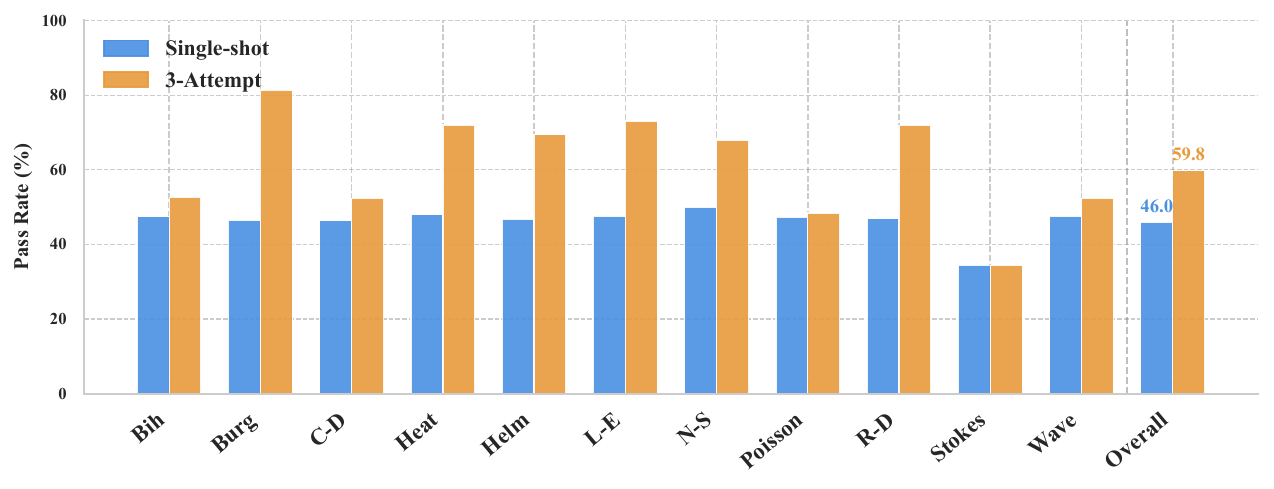}
  \caption{DOLFINx }
  \label{fig:iterative_DOLFINx}
\end{subfigure}
\hfill
\begin{subfigure}[b]{0.30\textwidth}
  \centering
  \includegraphics[width=\linewidth]{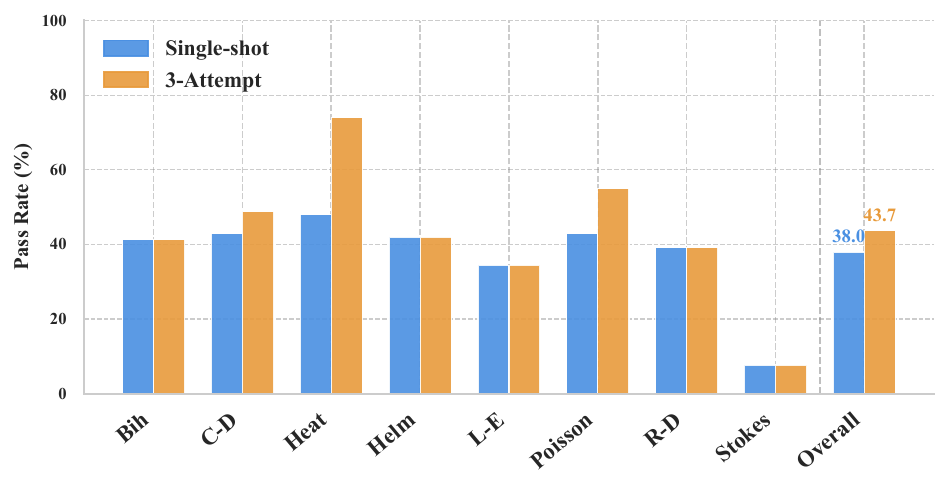}
  \caption{Firedrake}
  \label{fig:iterative_firedrake}
\end{subfigure}
\hfill
\begin{subfigure}[b]{0.30\textwidth}
  \centering
  \includegraphics[width=\linewidth]{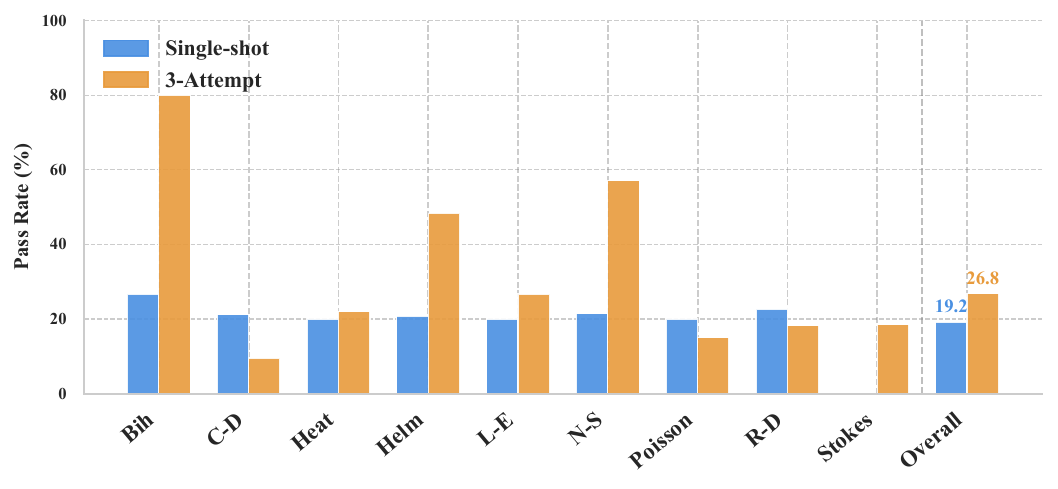}
  \caption{deal.II}
  \label{fig:iterative_dealii}
\end{subfigure}
\caption{GPT-5.4 pass rates under single-shot and three-attempt execution-feedback settings.}
\label{fig:iterative_comparison}
\end{figure}

Figure~\ref{fig:iterative_comparison} compares GPT-5.4 under the single-shot setting and the three-attempt execution-feedback setting. Execution feedback improves pass rates on all three FEM-library tracks. The largest gain appears on DOLFINx, where the pass rate increases from $46.0\%$ to $59.8\%$. Firedrake improves from $38.0\%$ to $43.7\%$, and deal.II improves from $19.2\%$ to $26.8\%$. These gains are accompanied by improved execution validity, indicating that a substantial fraction of single-shot failures arise from repairable compilation errors, runtime errors, output-format issues, or library API mistakes.

At the PDE-family level, gains are broadest on DOLFINx but more limited on Firedrake and deal.II. 
Stokes problems, transport-dominated cases, and the C++ implementation complexity of deal.II remain difficult to repair within a few feedback rounds. 
Overall, execution feedback mainly mitigates implementation-level errors, but remains limited for failures requiring correct numerical formulation, discretization choices, or solver configuration.

\subsection{Template-Guided Numerical Reasoning}
\label{sec:template_ablation}

The default end-to-end setting requires a model to perform two tasks simultaneously: making numerical decisions for a given PDE and implementing a complete solver in the target FEM library. To reduce the confounding effect of library boilerplate and API usage, we introduce a template-guided ablation in which the model receives a runnable solver skeleton and fills in the variational form, boundary or initial conditions, discretization choices, and solver parameters. This setting preserves the core numerical reasoning requirements while partially controlling for library-specific implementation burden. Appendix~\ref{app:template_details} provides additional details.

\begin{figure}[H]
\centering
\begin{subfigure}[b]{0.32\textwidth}
  \centering
  \includegraphics[width=\linewidth]{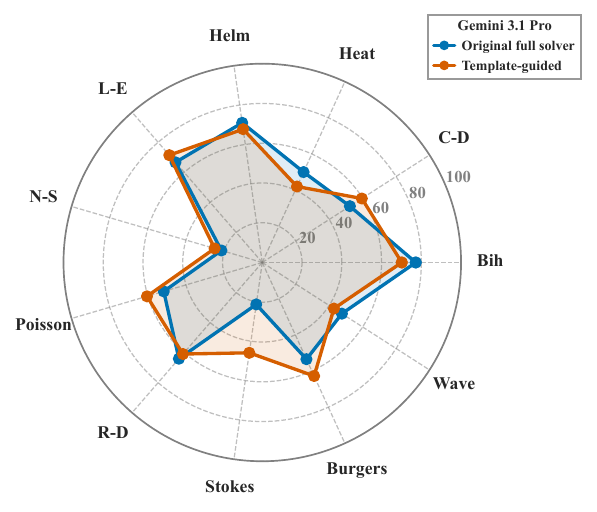}
  \caption{DOLFINx}
  \label{fig:template_ablation_DOLFINx}
\end{subfigure}
\hfill
\begin{subfigure}[b]{0.32\textwidth}
  \centering
  \includegraphics[width=\linewidth]{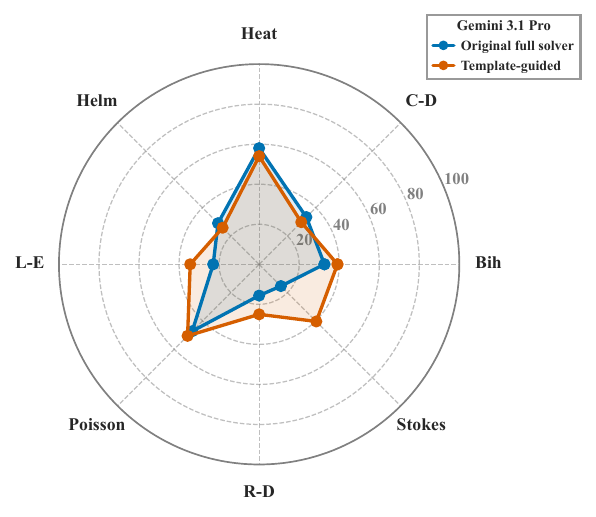}
  \caption{Firedrake}
  \label{fig:template_ablation_firedrake}
\end{subfigure}
\hfill
\begin{subfigure}[b]{0.32\textwidth}
  \centering
  \includegraphics[width=\linewidth]{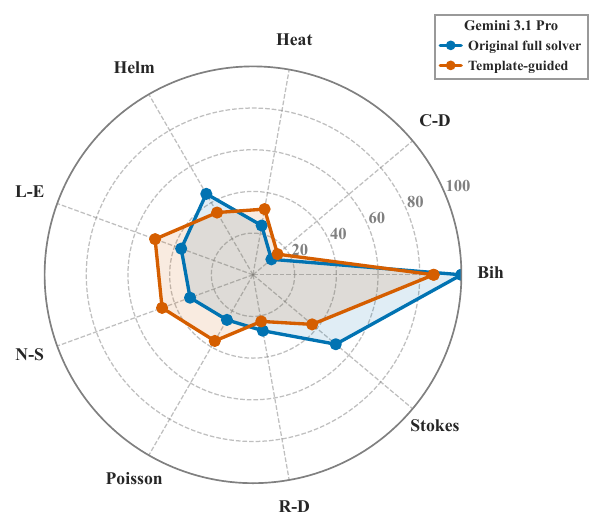}
  \caption{deal.II}
  \label{fig:template_ablation_dealii}
\end{subfigure}
\caption{Template-guided ablation for Gemini 3.1 Pro, comparing default end-to-end generation with template-guided generation.}

\label{fig:template_ablation}
\end{figure}
\vspace{-0.2cm}

Figure~\ref{fig:template_ablation} compares Gemini 3.1 Pro under the default end-to-end setting and the template-guided setting. The template yields consistent but modest gains: DOLFINx improves from $54.1\%$ to $56.3\%$, Firedrake from $32.4\%$ to $35.3\%$, and deal.II from $32.3\%$ to $35.8\%$. These results show that library boilerplate and API details are measurable sources of failure, but they do not fully explain the remaining performance gap. Some formulation-heavy PDE families improve under templating, while others decline, suggesting that a fixed scaffold can also constrain otherwise viable implementations. Thus, the template-guided track is best viewed as a diagnostic ablation: it reduces part of the implementation burden, but remaining failures still reflect difficulty with weak forms, finite-element spaces, boundary-condition realization, time stepping, and solver configuration.

\section{Discussion and Limitations}
\label{sec:discussion}

PDEAgent-Bench is a large-scale benchmark for PDE-to-solver synthesis that evaluates whether LLMs and code agents can produce solvers that are executable, accurate, and efficient. Unlike prior benchmarks, it uses a staged evaluation—checking execution, numerical accuracy, and runtime—since assessing only executability can overestimate solver quality.

The benchmark still has limitations. For instance, the dataset is primarily constructed using manufactured solutions, or, when analytical solutions are unavailable, high-accuracy numerical solutions are used as references. This design ensures reproducible, automated evaluation, but it may not fully reflect the complexity and diversity of PDE modeling in real-world industrial or engineering applications~\cite{huang2025self}, such as large-scale 3D fluid-structure simulations, multi-condition scenarios, and multiphysics couplings. Notably, the development of PDE benchmarks shows a progression from early, purely simulation-based PDEBench datasets~\citep{takamoto2022pdebench, gupta2022towards, luo2024cfdbench, wang2024scibench, zhang2024scicode} to more realistic settings in RealPDEBench~\citep{hu2026realpdebench}. In contrast, PDE code agent~\citep{li2025codepde, wu2025automatedcodedevelopmentpde, cheng2026re4scientificcomputingagent, soroco2025pdecontroller} research has yet to move beyond generating executable solvers, and our work provides a systematic evaluation platform for this foundational step, establishing a benchmark for future extensions toward real scientific and engineering tasks.

\section{Conclusion}
\label{sec:conclusion}

We introduce PDEAgent-Bench, the first multi-metric, multi-library benchmark for PDE-to-solver code generation by LLMs and code agents. It contains 645 PDE instances across 11 families and three FEM-library tracks: DOLFINx, Firedrake, and deal.II. The benchmark uses a staged evaluation—checking execution, numerical accuracy, and runtime. Experiments show that while strong models can produce runnable solvers for many instances, performance drops under accuracy, runtime, cross-library transfer, and challenging PDEs. PDEAgent-Bench provides a reproducible, diagnostic, and extensible foundation for evaluating pde code agents not just for code correctness, but for numerical accuracy and computational performance.

\bibliography{icml2026}
\bibliographystyle{unsrtnat}

\clearpage
\appendix

\section{Related Work}

\subsection{Code Generation Benchmarks and Code Agents}

Code generation benchmarks have become standard for evaluating code agents, often using execution-based evaluation to assess agent performance. For example, SWE-bench \citep{jimenez2024swebench} formulates tasks as patch generation for real-world issues and verifies correctness through test suites.
Recent benchmarks such as SWE-Bench Pro \citep{deng2025swebenchproaiagents}, GitTaskBench \citep{ni2025gittaskbenchbenchmarkcodeagents}, DA-Code \citep{huang-etal-2024-da}, and RACE-Bench \citep{liu2026benchmarkevaluatingrepositorylevelcode} further cover long-horizon software engineering, repository-level task solving, data-science coding workflows, and intermediate reasoning. 
Beyond pure coding, generic tool-use and interactive environment benchmarks---including AgentBench \citep{liu2024agentbench}, WebArena \citep{zhou2024webarena}, OSWorld \citep{xie2024osworld}, and MCP-Bench \citep{wang2026mcpbench}---evaluate models across diverse execution contexts. 
As these tasks grow in complexity, platforms like ML-Bench \citep{tang2026mlbench} and ResearchCodeBench \citep{hua2025researchcodebench} have been proposed to rigorously evaluate whether agents can execute repository-level ML workflows or implement genuinely novel research ideas. Similarly, MLR-Bench \citep{chen2026mlrbench} and HAL \citep{ndzomga2026efficientbenchmarkingaiagents} address the end-to-end evaluation of open-ended scientific research and the efficiency of benchmarking itself.
Building on this paradigm, agent systems such as SWE-agent \citep{yang2024sweagent} and OpenHands \citep{wang2024openhands} incorporate planning, editing, and iterative execution to handle more complex engineering tasks.
While these benchmarks are well suited for assessing discrete functional correctness in conventional software development or ML scripting, PDE solver synthesis further requires generated programs to satisfy strict numerical accuracy and physics-based runtime constraints.

\subsection{PDE Benchmarks and PDE Code Agents}

PDE benchmarks in scientific machine learning, such as PDEBench, PDEArena, and CFDBench \citep{takamoto2022pdebench, gupta2022towards, luo2024cfdbench}, primarily evaluate the predictive accuracy and generalization of learned surrogate solution operators, rather than the synthesis of executable PDE solvers from problem specifications.
Other scientific benchmarks cover adjacent capabilities: SciBench \citep{wang2024scibench} and SciCode \citep{zhang2024scicode} evaluate scientific reasoning and scientific scripting, PHYBench \citep{liu2025phybench} focuses on physical reasoning, and PINNacle \citep{hao2024pinnacle} provides a systematic assessment of physics-informed neural networks.

Driven by the broader rise of scientific intelligence \citep{ren2026scientificintelligencesurveyllmbased}, recent work explores LLM-based agents for diverse scientific domains, spanning autonomous materials discovery \citep{ruza2026reasoningtosimulation} and foundational zero-shot PDE extrapolation \citep{bao2026texttrainedllmszeroshotextrapolate}.
Recent advances also highlight the potential of fusing multimodal text descriptions directly into PDE neural operators \citep{lorsung2025explain} and using LLM agents to autonomously automate the hyperparameter search and architecture optimization of PDE surrogates \citep{wuwu2025pinnsagent}.
Specifically for solver code generation, CodePDE \citep{li2025codepde} studies LLM-driven synthesis; LLM-PDEDeveloper and Re4 \citep{wu2025automatedcodedevelopmentpde, cheng2026re4scientificcomputingagent} investigate agentic workflows for scientific computing code generation; and PDE-Controller, LLM4PD, and OpInf-LLM \citep{soroco2025pdecontroller, luo-etal-2025-large-language, wang2026opinfllm} examine related capabilities in PDE autoformalization, PDE discovery, and parametric PDE solving.
However, these works are typically evaluated on limited or heterogeneous task collections, making unified and reproducible comparison difficult.
Overall, existing work lacks a multi-library, staged benchmark for PDE-to-solver code generation that jointly evaluates executability, numerical accuracy, and computational efficiency.

\section{Cross-Library Per-Family Results}
\label{app:eqtype_firedrake_dealii}

Table~\ref{tab:eqtype_firedrake_dealii} reports single-shot case-level pass rates for all models on the Firedrake and deal.II, broken down by PDE equation family. These results complement the DOLFINx primary table (Table~\ref{tab:eqtype_backend_pass_rates}) and the per-family visualisation in Figure~\ref{fig:backend_eqtype}.

\begin{table*}[h]
\centering
\scriptsize
\setlength{\tabcolsep}{4pt}
\renewcommand{\arraystretch}{1.25}
\newcommand{\rh}[1]{\rotatebox{55}{\textbf{#1}}}
\resizebox{\textwidth}{!}{%
\begin{tabular}{@{} ll l *{11}{c} >{\columncolor{allfill}}c @{}}
\toprule
\rowcolor{hdrfill}
\textbf{FEM library} & \textbf{Type} & \textbf{Model} &
\rh{Biharmonic} & \rh{Conv.-Diff.} & \rh{Heat} & \rh{Helmholtz} &
\rh{Lin.\ Elast.} & \rh{Stokes} & \rh{Poisson} &
\rh{React.-Diff.} & \rh{N.-Stokes} & \rh{Burgers} & \rh{Wave} &
\textbf{All} \\
\midrule

\multirow{9}{*}{\textbf{Firedrake}}
& \multirow{6}{*}{\makecell[l]{Base\\LLM}}
& GPT-5.4        & 41.3 & 42.9 & 48.0 & 41.9 & 34.4 &  7.7 & 42.9 & 39.1 & -- & -- & -- & 38.0 \\
& & Opus 4.7       & 19.6 & \textbf{47.6} & \textbf{76.0} & \textbf{40.3} & \textbf{44.3} & \textbf{25.0} & \textbf{54.9} & \textbf{50.0} & -- & -- & -- & \textbf{45.9} \\
& & Gemini 3.1 Pro & \textbf{32.6} & 33.3 & 58.0 & 29.0 & 23.0 & 15.4 & 47.3 & 15.6 & -- & -- & -- & 32.4 \\
& & Qwen3.6-Plus   &  32.6 &  25.0 &  26.0 &  17.7 &  23.0 &  15.4 &  27.5 &  15.6 & -- & -- & -- &  22.9 \\
& & DeepSeek V3.2  &  2.2 &  0.0 &  0.0 &  0.0 &  0.0 &  0.0 & 14.3 &  0.0 & -- & -- & -- &  2.7 \\
\cmidrule(l){2-15}
& \multirow{3}{*}{\makecell[l]{Code/PDE\\Agent}}
& CodePDE        & 10.9 & 26.2 & 56.0 & 21.0 &  9.8 &  7.7 & 37.4 & 26.6 & -- & -- & -- & 25.3 \\
& & OpenHands     & 19.6 & 39.3 & 42.0 & 32.3 &  9.8 &  7.7 & 36.3 & 32.8 & -- & -- & -- & 28.8 \\
& & Mini-SWE      &  8.7 & 27.4 & 12.0 &  8.1 &  4.9 &  3.8 & 18.7 &  9.4 & -- & -- & -- & 12.9 \\

\specialrule{1.2pt}{2pt}{2pt}

\multirow{9}{*}{\textbf{deal.II}}
& \multirow{6}{*}{\makecell[l]{Base\\LLM}}
& GPT-5.4        & 26.7 & 21.2 & 20.0 & 20.7 & 20.0 &  0.0 & 20.0 & 22.7 & 21.4 & -- & -- & 19.2 \\
& & Opus 4.7       & 93.3 & \textbf{15.4} &  0.0 & \textbf{44.8} & \textbf{40.0} & 44.4 & \textbf{28.3} &  0.0 & \textbf{42.9} & -- & -- & 28.1 \\
& & Gemini 3.1 Pro & \textbf{100.0} & 11.5 & \textbf{24.0} & \textbf{44.8} & 36.7 & \textbf{51.9} & 25.0 & \textbf{27.3} & 32.1 & -- & -- & \textbf{32.3} \\
& & Qwen3.6-Plus   & 33.3 &  15.4 &  12.0 &  10.3 &  23.3 &  14.8 &  8.3 &  27.3 &  32.1 & -- & -- &  16.9 \\
& & DeepSeek V3.2  &  0.0 &  0.0 &  0.0 &  0.0 &  0.0 &  0.0 &  0.0 &  0.0 &  0.0 & -- & -- &  0.0 \\
\cmidrule(l){2-15}
& \multirow{3}{*}{\makecell[l]{Code/PDE\\Agent}}
& CodePDE        & 33.3 & 15.4 & 22.0 & 31.0 & 33.3 & 14.8 & 25.0 & 27.3 & 32.1 & -- & -- & 24.6 \\
& & OpenHands     & 46.7 &  9.6 &  6.0 & 31.0 & 13.3 &  7.4 & 16.7 & 13.6 & 32.1 & -- & -- & 16.6 \\
& & Mini-SWE      &  0.0 &  0.0 &  0.0 &  0.0 &  0.0 &  0.0 &  0.0 &  0.0 &  0.0 & -- & -- &  0.0 \\

\bottomrule
\end{tabular}%
}
\caption{Single-shot case-level pass rates (\%) by PDE family on the Firedrake and deal.II cross-library tracks. Dashes indicate PDE families not included in the corresponding library track. Bold indicates the best result within each library--family column.}
\label{tab:eqtype_firedrake_dealii}
\end{table*}

Table~\ref{tab:eqtype_firedrake_dealii} reports single-shot case-level pass rates on the Firedrake and deal.II tracks, broken down by PDE family and model. These results complement the primary DOLFINx results in Table~\ref{tab:eqtype_backend_pass_rates} and the cross-library visualization in Figure~\ref{fig:backend_eqtype}. While the main text summarizes aggregate trends, this table shows that the cross-library degradation is not uniform across PDE families.

Several patterns are visible. On Firedrake, the strongest overall model is Opus 4.7, with an overall pass rate of 45.9\%, followed by GPT-5.4 at 38.0\% and Gemini 3.1 Pro at 32.4\%. The family-level results show that Firedrake performance remains relatively strong on Heat, Poisson, Convection--Diffusion, and Reaction--Diffusion, but drops sharply on Stokes for all systems. This suggests that the Firedrake track is not merely harder because of API differences; mixed finite-element formulations, saddle-point structure, and pressure-space handling remain persistent sources of failure.

On deal.II, overall pass rates are lower than on DOLFINx and Firedrake, reflecting the additional burden of generating correct and efficient C++ FEM implementations. Gemini 3.1 Pro obtains the highest overall pass rate at 32.3\%, followed by Opus 4.7 at 28.1\% and CodePDE at 24.6\%. However, the family-level distribution is highly uneven. Biharmonic cases are solved at high rates by Opus 4.7 and Gemini 3.1 Pro, whereas Stokes, Heat, and several transport or reaction families remain difficult for most models. These results indicate that cross-library transfer failures are family dependent: some models can reproduce isolated formulations in deal.II, but robust solver synthesis across coupled, time-dependent, or stability-sensitive PDEs remains limited.

\section{Dataset Documentation and FEM-Library Coverage}
\label{app:dataset_documentation}

Table~\ref{tab:dataset_overview} summarizes the composition of PDEAgent-Bench and the FEM-library coverage of each PDE family. The benchmark contains 645 PDE instances across 11 equation families. DOLFINx provides full coverage over all instances and serves as the primary track, while Firedrake and deal.II provide cross-library tracks with broad but partial coverage. Firedrake covers 510 instances across 8 PDE families, and deal.II covers 313 instances across 9 PDE families. This design supports both full-track evaluation on a primary FEM stack and cross-library evaluation across alternative Python and C++ finite-element ecosystems.

\begin{table*}[h]
\centering
\scriptsize
\caption{Dataset composition and FEM library-track coverage across PDE families.}
\label{tab:dataset_overview}
\begin{subtable}[t]{0.49\textwidth}
\centering
\caption{Dataset distribution across PDE families (total 645 instances).}
\label{tab:dataset_distribution}
\setlength{\tabcolsep}{3pt}
\begin{tabularx}{\linewidth}{>{\raggedright\arraybackslash}X>{\raggedright\arraybackslash}Xr}
\toprule
\textbf{PDE Family} & \textbf{Math Type} & \textbf{Instances} \\
\midrule
Poisson & elliptic & 91 \\
Helmholtz & elliptic & 62 \\
Biharmonic & elliptic & 57 \\
Linear Elasticity & elliptic & 63 \\
Heat & parabolic & 50 \\
Convection--Diffusion & elliptic/mixed type/parabolic & 84 \\
Reaction--Diffusion & reaction diffusion & 64 \\
Stokes & elliptic/incompressible flow & 61 \\
Navier--Stokes & incompressible flow & 28 \\
Burgers & parabolic & 43 \\
Wave & hyperbolic & 42 \\
\midrule
\textbf{Total} &  & \textbf{645} \\
\bottomrule
\end{tabularx}
\end{subtable}\hfill
\begin{subtable}[t]{0.49\textwidth}
\centering
\caption{Per-family FEM library-track coverage. Each cell shows the number of supported instances.}
\label{tab:library_support}
\setlength{\tabcolsep}{2pt}
\begin{tabularx}{\linewidth}{>{\raggedright\arraybackslash}Xccc}
\toprule
\textbf{PDE Family} & \textbf{DOLFINx} & \textbf{Firedrake} & \textbf{deal.II} \\
\midrule
Poisson & 91 & 91 & 60 \\
Helmholtz & 62 & 62 & 29 \\
Biharmonic & 57 & 46 & 15 \\
Linear Elasticity & 63 & 61 & 30 \\
Heat & 50 & 50 & 50 \\
Conv.--Diff. & 84 & 84 & 52 \\
React.--Diff. & 64 & 64 & 22 \\
Stokes & 61 & 52 & 27 \\
Navier--Stokes & 28 & --- & 28 \\
Burgers & 43 & --- & --- \\
Wave & 42 & --- & --- \\
\midrule
\textbf{Total} & \textbf{645} & \textbf{510} & \textbf{313} \\
\bottomrule
\end{tabularx}
\end{subtable}
\end{table*}

We provide a Datasheet for Datasets~\cite{gebru2021datasheets} to document the motivation, composition, collection process, intended uses, distribution, and maintenance of PDEAgent-Bench. The full datasheet is released alongside the benchmark; here we summarize the principal entries most relevant to reproducibility and responsible use.

\paragraph{Motivation.}
PDEAgent-Bench was created to evaluate end-to-end PDE-to-solver code generation. It addresses the lack of benchmarks that simultaneously assess executability, numerical accuracy, computational efficiency, and proficiency with multiple professional finite-element libraries. The dataset was curated by the authors as an academic research artifact; no third-party funding influenced case selection.

\paragraph{Composition.}
The dataset consists of $645$ PDE \emph{cases} grouped into $11$ equation families: Poisson, Helmholtz, Biharmonic, Linear Elasticity, Heat, Convection--Diffusion, Reaction--Diffusion, Stokes, Navier--Stokes, Burgers, and Wave. Cases are labeled with mathematical categories drawn from \texttt{elliptic}, \texttt{parabolic}, \texttt{hyperbolic}, \texttt{mixed\_type}, \texttt{incompressible\_flow}, and \texttt{reaction\_diffusion}. They span $14$ geometric templates: unit square, unit cube, L-shape, square-with-hole, multi-hole, circle, annulus, eccentric annulus, sector, star, gear, T-junction, dumbbell, and periodic square. Each case is a self-contained record containing the PDE specification, evaluation grid, calibration-baseline configuration, evaluator-side reference metadata, and supported FEM-library tracks among \{DOLFINx, Firedrake, deal.II\}. The released agent-facing task records do \emph{not} contain executable solver code generated by agents.

\paragraph{Collection process.}
All cases were synthetically constructed by the authors. PDE coefficients, source terms, boundary conditions, and initial conditions are derived analytically from manufactured solutions where possible. For cases without analytic manufactured solutions, a higher-fidelity numerical reference is generated using the reference-generation pipeline with refined meshes, higher-order discretizations, and stricter solver tolerances. No personal, proprietary, or otherwise restricted data is included.

\paragraph{Agent-visible and evaluator-only fields.}
Each instance separates agent-visible information from evaluator-only metadata. The agent-visible fields include the PDE family, domain description, coefficient functions, source terms, boundary and initial conditions, temporal parameters when applicable, and the required output-grid specification. Evaluator-only fields include reference solutions, calibration baselines, reference-generation settings, threshold-setting metadata, and hidden numerical configurations. Analytic manufactured solutions, when used to construct source terms or reference values, are retained only in evaluator-side metadata and are not exposed as target solution fields in the agent prompt.

\paragraph{Preprocessing, cleaning, and labeling.}
Each candidate record passes a static validation suite checking schema conformance, identifier uniqueness, cross-field consistency, symbolic parsability of expressions, boundary-tag validity, and the presence of a high-fidelity \texttt{reference\_config} when no manufactured solution is available. Records are then trial-executed in the evaluation harness to verify reference generation, calibration execution, output-grid sampling, format checking, and stable error computation. Cases that fail any validation or trial-execution check are discarded and not released.

\paragraph{Uses.}
The intended use is benchmarking LLM-based code agents and code generation systems on PDE-to-solver tasks. PDEAgent-Bench is designed for research on scientific code generation, numerical reasoning, FEM-library usage, and agent evaluation. Inappropriate uses include directly deploying LLM-generated solvers in safety-critical engineering or scientific pipelines without independent verification. Passing PDEAgent-Bench should be interpreted as benchmark performance, not as engineering qualification.

\paragraph{Distribution and license.}
The case records are released under a permissive license suitable for benchmark redistribution. The evaluation harness, reference-generation scripts, and Docker recipes are released under an OSI-approved open-source license. During review, the anonymized repository includes the case records, evaluator, container recipes, prompts, and baseline outputs needed to reproduce the reported results. Persistent identifiers, the leaderboard portal, and a long-term archival mirror will be provided in the camera-ready version.

\paragraph{Maintenance.}
The benchmark is maintained by the corresponding author group. Versioning follows semantic versioning, and changes are tracked in a public changelog. Bug reports and contribution proposals are accepted through the project issue tracker. Prior major versions are retained for at least two years to support reproducibility of published results.

\section{Additional Solution-Quality and Solver-Behavior Diagnostics}
\label{app:additional_diagnostics}

The staged evaluation protocol reports whether a generated solver executes, satisfies the calibrated accuracy threshold, and runs within the runtime budget. In this appendix, we provide additional metrics over passed cases to characterize the numerical quality and solver behavior behind the binary pass labels. These metrics are diagnostic only and do not affect the benchmark pass/fail decision.

\begin{table*}[t]
\centering
\scriptsize
\setlength{\tabcolsep}{3pt}
\resizebox{\textwidth}{!}{%
\renewcommand{\arraystretch}{1.4}
\begin{tabular}{l|cccc|cccc|cccc}
\toprule
\multicolumn{1}{c|}{} &
\multicolumn{4}{c|}{\textbf{DOLFINx}} &
\multicolumn{4}{c|}{\textbf{Firedrake}} &
\multicolumn{4}{c}{\textbf{deal.II}} \\
\cmidrule(lr){2-5}\cmidrule(lr){6-9}\cmidrule(lr){10-13}
Model &
RMSE$\downarrow$ & MAE$\downarrow$ & $R^2\uparrow$ & fRMSE$\downarrow$ &
RMSE$\downarrow$ & MAE$\downarrow$ & $R^2\uparrow$ & fRMSE$\downarrow$ &
RMSE$\downarrow$ & MAE$\downarrow$ & $R^2\uparrow$ & fRMSE$\downarrow$ \\
\midrule
GPT-5.4
& $2.16{\times}10^{-3}$ & $1.51{\times}10^{-3}$ & 0.997 & $2.15{\times}10^{-3}$
& $1.42{\times}10^{-3}$ & $1.05{\times}10^{-3}$ & 0.998 & $1.42{\times}10^{-3}$
& $5.43{\times}10^{-4}$ & $\underline{6.12{\times}10^{-5}}$ & 0.752 & $7.34{\times}10^{-4}$ \\

Opus 4.7
& $\mathbf{2.16{\times}10^{-4}}$ & $\mathbf{1.69{\times}10^{-4}}$ & \underline{0.999} & $\mathbf{2.03{\times}10^{-4}}$
& $\underline{5.88{\times}10^{-4}}$ & $\underline{4.42{\times}10^{-4}}$ & 0.998 & $\underline{5.86{\times}10^{-4}}$
& $\underline{1.00{\times}10^{-4}}$ & $7.93{\times}10^{-5}$ & \textbf{1.000} & $\underline{1.00{\times}10^{-4}}$ \\

Gemini 3.1 Pro
& $\underline{2.76{\times}10^{-4}}$ & $\underline{2.22{\times}10^{-4}}$ & \underline{0.999} & $\underline{2.46{\times}10^{-4}}$
& $1.36{\times}10^{-2}$ & $1.01{\times}10^{-2}$ & 0.942 & $1.20{\times}10^{-2}$
& $\mathbf{8.23{\times}10^{-5}}$ & $\mathbf{5.79{\times}10^{-5}}$ & \textbf{1.000} & $\mathbf{8.23{\times}10^{-5}}$ \\

Qwen3.6-Plus
& $7.26{\times}10^{-4}$ & $6.01{\times}10^{-4}$ & \textbf{1.000} & $7.26{\times}10^{-4}$
& $1.09{\times}10^{-2}$ & $3.12{\times}10^{-3}$ & 0.823 & $9.80{\times}10^{-3}$
& $7.07{\times}10^{-2}$ & $5.57{\times}10^{-2}$ & \underline{0.920} & $7.07{\times}10^{-2}$ \\

DeepSeek V3.2
& $1.42{\times}10^{-3}$ & $1.07{\times}10^{-3}$ & 0.987 & $1.42{\times}10^{-3}$
& $\mathbf{1.25{\times}10^{-5}}$ & $\mathbf{8.44{\times}10^{-6}}$ & \textbf{1.000} & $\mathbf{1.25{\times}10^{-5}}$
& -- & -- & -- & -- \\
\midrule
CodePDE
& $2.09{\times}10^{-3}$ & $1.57{\times}10^{-3}$ & 0.998 & $2.01{\times}10^{-3}$
& $1.25{\times}10^{-3}$ & $9.19{\times}10^{-4}$ & 0.998 & $1.21{\times}10^{-3}$
& $7.33{\times}10^{-2}$ & $5.24{\times}10^{-2}$ & 0.739 & $7.33{\times}10^{-2}$ \\

OpenHands
& $1.16{\times}10^{-3}$ & $8.94{\times}10^{-4}$ & 0.998 & $1.12{\times}10^{-3}$
& $1.93{\times}10^{-3}$ & $1.53{\times}10^{-3}$ & 0.998 & $1.93{\times}10^{-3}$
& $1.15{\times}10^{-2}$ & $9.14{\times}10^{-3}$ & 0.916 & $1.15{\times}10^{-2}$ \\

Mini-SWE
& $5.63{\times}10^{-4}$ & $4.01{\times}10^{-4}$ & \underline{0.999} & $5.43{\times}10^{-4}$
& $2.08{\times}10^{-3}$ & $1.71{\times}10^{-3}$ & \underline{0.999} & $2.08{\times}10^{-3}$
& -- & -- & -- & -- \\
\bottomrule
\end{tabular}%
}
\caption{Universal solution-quality metrics averaged over all passed cases, by model and FEM library. ``--'' indicates that the corresponding model--library pair has no passed cases for which the metric is reported. \textbf{Bold}: best per column; \underline{underline}: second best per column.}
\label{tab:metrics_backends}
\end{table*}

\paragraph{Solution quality on passing cases.}
Table~\ref{tab:metrics_backends} reports solution-quality metrics averaged over passed cases for each model and FEM library. These numbers should be interpreted as conditional solution quality rather than a common-case ranking, because each model is averaged over its own set of passed instances. Even with this qualification, the results show that binary pass labels can conceal substantial variation in numerical precision. On DOLFINx, Opus 4.7 and Gemini 3.1 Pro achieve $R^2 \geq 0.999$ with RMSE in the $10^{-4}$ range, whereas GPT-5.4 has $\mathrm{RMSE}=2.16{\times}10^{-3}$, suggesting a larger share of threshold-marginal passes. On deal.II, CodePDE has $R^2=0.739$ and $\mathrm{RMSE}=7.33{\times}10^{-2}$ over its passed cases, indicating that some solvers can clear the calibrated accuracy gate while remaining numerically marginal. The frequency-domain fRMSE generally tracks spatial RMSE, providing a complementary view of error structure.

\begin{table*}[t]
\centering
\scriptsize
\setlength{\tabcolsep}{2.3pt}
\resizebox{\textwidth}{!}{%
\renewcommand{\arraystretch}{1.4}
\begin{tabular}{l|cccc|cccc|ccc|ccc|ccc|ccc}
\toprule
\multicolumn{1}{c|}{} &
\multicolumn{4}{c|}{\textbf{Elliptic}} &
\multicolumn{4}{c|}{\textbf{Parabolic}} &
\multicolumn{3}{c|}{\textbf{Hyperbolic}} &
\multicolumn{3}{c|}{\textbf{Incom. flow}} &
\multicolumn{3}{c|}{\textbf{Reaction-diffusion}} &
\multicolumn{3}{c}{\textbf{Mixed type}} \\
\cmidrule(lr){2-5}\cmidrule(lr){6-9}\cmidrule(lr){10-12}
\cmidrule(lr){13-15}\cmidrule(lr){16-18}\cmidrule(lr){19-21}
Model &
Eff. & DOF & Res. & Lin.Iter &
WorkRate & CFL & Time/s & Lin.Iter &
Res. & Deg. & Lin.Iter &
Res. & Lin.Iter & Newton &
Res. & Lin.Iter & Newton &
Pe & Lin.Iter & TV \\
\midrule
GPT-5.4
& 26104 & 32108 & 92.8 & 131.6
& 53289 & 45.7 & 0.097 & 55.7
& 66.7 & 1.0 & 227.3
& 64.4 & 0.3 & 0.3
& 82.4 & 401.8 & 3.5
& 3317 & 223 & 123.4 \\

Opus 4.7
& 15332 & 29317 & 73.1 & 29.8
& 53288 & 51.0 & 0.253 & 127.0
& 62.6 & 1.4 & 227.2
& 53.2 & 1.1 & 3.1
& 66.7 & 85.9 & 1.9
& 3100 & 97.2 & 199.8 \\

Gemini 3.1 Pro
& 23262 & 47766 & 86.9 & 41.2
& 154774 & 76.4 & 0.220 & 990.3
& 68.5 & 2.0 & 474.8
& 68.0 & 1.0 & 4.3
& 54.7 & 642.3 & 2.0
& 3381 & 240 & 123.2 \\

Qwen3.6-Plus
& 23187 & 28892 & 100.6 & 65.6
& 38582 & 163.8 & 0.425 & 763.0
& 73.1 & 1.8 & 328.1
& 72.1 & 3.8 & 2.3
& 67.4 & 174.5 & 2.3
& 2789 & 198 & 119.2 \\

DeepSeek V3.2
& 96488 & 172855 & 152.7 & 36.8
& 25805 & 82.8 & 0.391 & 35.3
& -- & -- & --
& 80.0 & 7.0 & 1.8
& 84.3 & 50.0 & 2.2
& -- & -- & -- \\

CodePDE
& 20024 & 41701 & 103.8 & 82.4
& 21769 & 72.0 & 0.424 & 82.1
& 88.4 & 1.0 & 255.5
& 65.1 & 0.6 & 4.2
& 96.4 & 221.4 & 2.35
& 3246 & 187 & 119.8 \\

OpenHands
& 73023 & 53836 & 87.1 & 173.8
& 159314 & 369.8 & 0.269 & 79.5
& 88.8 & 1.4 & 165.2
& 76.2 & 2.9 & 1.9
& 82.8 & 452.9 & 1.95
& 2378 & 50.9 & 233.1 \\

Mini-SWE
& 12850 & 18551 & 58.8 & 98.3
& 2591973 & 6804.2 & 0.119 & 36.7
& 80.0 & 1.0 & 280.0
& 84.0 & 0.5 & 0.0
& 88.0 & 430.0 & 2.00
& 1118 & 24 & 1.0 \\
\bottomrule
\end{tabular}%
}
\caption{Numerical diagnostic metrics by mathematical problem type, averaged over passed DOLFINx cases. The table summarizes discretization scale, solver behavior, and runtime efficiency. ``--'' indicates no passed cases for the corresponding math type.}
\label{tab:metrics_all_regimes}
\end{table*}

\paragraph{Solver-behavior diagnostics.}
Table~\ref{tab:metrics_all_regimes} reports numerical-behavior diagnostics over passed DOLFINx cases, including discretization scale, CFL values, linear-iteration counts, nonlinear-iteration counts, residual indicators, and work-rate statistics. These diagnostics show that the same binary pass label can correspond to qualitatively different solver behavior. For example, Gemini 3.1 Pro obtains many elliptic passes with moderate mesh sizes and linear-iteration counts, whereas OpenHands attains many parabolic passes with substantially larger average CFL values, suggesting more aggressive time-step choices or different stability assumptions. DeepSeek V3.2's few elliptic passes rely on unusually large meshes, consistent with a strategy closer to brute-force refinement. These results support the use of multi-metric reporting: pass rate measures benchmark success, while the diagnostic metrics reveal how generated solvers achieve that success.

\section{Threshold Sensitivity Analysis}
\label{app:threshold_sensitivity}

The evaluation framework in Section~\ref{sec:scoring_leaderboard} uses two global multipliers: $\alpha_{\mathrm{acc}} = 10$ for the accuracy gate and $\alpha_{\mathrm{time}} = 3$ for the runtime gate. These values are fixed for all models, PDE families, and instances. They are not tuned on model outcomes; instead, they are chosen to provide a transparent calibration-based criterion that permits reasonable implementation differences while still requiring numerically meaningful and computationally practical solvers.

This appendix analyzes the sensitivity of the DOLFINx primary track to both multipliers. We focus on DOLFINx because it is the full-coverage track containing all 645 benchmark instances. The Firedrake and deal.II tracks use the same thresholding protocol but have partial family coverage, as described in Appendix~\ref{app:dataset_documentation}. The goal of this analysis is not to select the most favorable threshold, but to verify that the main conclusions and model ordering are stable under reasonable variations of the accuracy and runtime gates.

\paragraph{Accuracy multiplier.}
The accuracy gate requires
\[
e(u_{\mathrm{agent}}) \leq \tau_{\mathrm{acc}},
\qquad
\tau_{\mathrm{acc}} = \max(\alpha_{\mathrm{acc}} e_{\mathrm{base}}, \tau_{\min}),
\]
where $e(u_{\mathrm{agent}})$ is the relative $L^2$ error of the submitted solution on the prescribed evaluation grid, $e_{\mathrm{base}}$ is the relative $L^2$ error of the calibration baseline, and $\tau_{\min}$ is a small accuracy floor used when the calibration error is near zero. The default value $\alpha_{\mathrm{acc}} = 10$ requires the submitted solution to be within one order of magnitude of the calibration-baseline error, subject to the minimum floor. This is deliberately permissive: agents are allowed to choose their own meshes, finite-element spaces, discretization orders, time-stepping schemes, and solver settings, and the benchmark should not require them to exactly reproduce the calibration solver.

The one-order-of-magnitude criterion also reflects the fact that numerically valid implementations can differ in discretization scale. For a method with empirical convergence order $q$, coarsening the mesh by a factor of two can increase the discretization error by approximately $2^q$, assuming the asymptotic regime and sufficient solution regularity. Thus, $\alpha_{\mathrm{acc}} = 10$ can accommodate modest mesh coarsening or lower-order implementation choices while still rejecting solutions that deviate substantially from the ground truth. Much smaller values, such as $\alpha_{\mathrm{acc}} \leq 2$, would penalize efficient but slightly coarser solvers; much larger values would increasingly accept qualitatively inaccurate solutions.

\begin{figure}[H]
\centering
\includegraphics[width=0.52\textwidth]{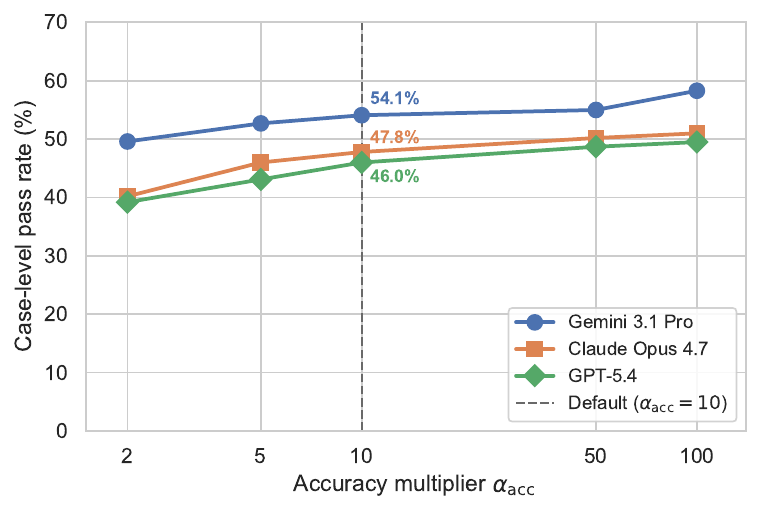}
\caption{DOLFINx case-level pass rate for the three top-performing models as a function of the accuracy multiplier $\alpha_{\mathrm{acc}}$, with $\alpha_{\mathrm{time}} = 3$ fixed. The DOLFINx track contains 645 instances. The dashed vertical line marks the default value $\alpha_{\mathrm{acc}} = 10$. Model rankings are stable across all tested values.}
\label{fig:acc_sensitivity}
\end{figure}

We evaluate sensitivity to the accuracy multiplier by varying
$\alpha_{\mathrm{acc}} \in \{2, 5, 10, 50, 100\}$ while holding $\alpha_{\mathrm{time}} = 3$ fixed. Figure~\ref{fig:acc_sensitivity} shows the DOLFINx case-level pass rates for the three strongest systems. The relative ordering is stable across the tested range: Gemini~3.1~Pro remains the top-performing model, followed by Opus~4.7 and GPT-5.4. Increasing $\alpha_{\mathrm{acc}}$ raises absolute pass rates, as expected, but the increase is gradual rather than discontinuous around the default value. This supports interpreting $\alpha_{\mathrm{acc}} = 10$ as a transparent same-order-of-magnitude accuracy criterion rather than a threshold selected to alter model rankings.

\paragraph{Runtime multiplier.}
The runtime gate requires:
\[
t_{\mathrm{agent}} \leq \tau_{\mathrm{time}},
\qquad
\tau_{\mathrm{time}} = \alpha_{\mathrm{time}} t_{\mathrm{base}},
\]
where $t_{\mathrm{agent}}$ is the wall-clock runtime of the submitted solver and $t_{\mathrm{base}}$ is the runtime of the calibration baseline under the same evaluation environment. The default value $\alpha_{\mathrm{time}} = 3$ allows a bounded overhead relative to the calibration baseline. This tolerance is intended to account for implementation differences, solver choices, and library-level overhead, while still excluding solutions that are impractically slow for the prescribed case.

\begin{figure}[H]
\centering
\includegraphics[width=0.52\textwidth]{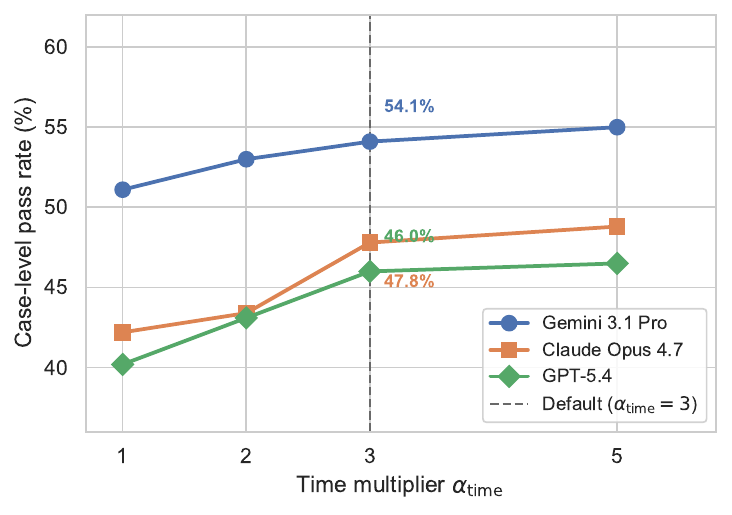}
\caption{DOLFINx case-level pass rate for the three top-performing models as a function of the runtime multiplier $\alpha_{\mathrm{time}}$, with $\alpha_{\mathrm{acc}} = 10$ fixed. The DOLFINx track contains 645 instances. The dashed vertical line marks the default value $\alpha_{\mathrm{time}} = 3$. Model rankings are stable across all tested values.}
\label{fig:time_sensitivity}
\end{figure}

We evaluate sensitivity to the runtime multiplier by varying
$\alpha_{\mathrm{time}} \in \{1, 2, 3, 5\}$ while holding $\alpha_{\mathrm{acc}} = 10$ fixed. Figure~\ref{fig:time_sensitivity} reports the resulting DOLFINx case-level pass rates for the three strongest systems. Across all tested values, the ordering Gemini~3.1~Pro $>$ Opus~4.7 $\approx$ GPT-5.4 is preserved. The absolute pass rates are also stable: varying $\alpha_{\mathrm{time}}$ from 1 to 5 changes each model's pass rate by at most 4 percentage points. This indicates that most submitted solvers either run comfortably within the runtime budget or fail by a wider margin, rather than clustering near the default threshold.

Overall, the sensitivity analysis shows that the main DOLFINx conclusions are robust to reasonable variations of both threshold multipliers. The default values $\alpha_{\mathrm{acc}} = 10$ and $\alpha_{\mathrm{time}} = 3$ provide a consistent calibrated evaluation rule: the accuracy gate requires same-order numerical agreement with the calibration baseline, and the runtime gate enforces a bounded efficiency requirement without materially changing the relative model ranking.

\section{Calibration Baseline Distribution and Reference-Solution Validity}
\label{app:calibration_distribution}

This appendix reports the empirical distribution of calibration-baseline errors and runtimes on the DOLFINx primary track, explains the role of the accuracy floor in Section~\ref{sec:scoring_leaderboard}, and describes how reference-solution credibility is checked for cases without analytic manufactured solutions. We focus on DOLFINx because it is the full-coverage track containing all 645 benchmark instances.

\paragraph{Reference solutions and calibration baselines.}
PDEAgent-Bench distinguishes the reference solution used for scoring from the calibration baseline used for threshold setting. For each case, the submitted solution $u_{\mathrm{agent}}$ is compared against a reference solution $u_{\mathrm{GT}}$ on the prescribed evaluation grid. Separately, a calibration baseline $u_{\mathrm{base}}$ is generated during benchmark construction and used to derive the case-specific accuracy threshold through
\[
\tau_{\mathrm{acc}} = \max(\alpha_{\mathrm{acc}} e_{\mathrm{base}}, \tau_{\min}),
\qquad
e_{\mathrm{base}} =
\frac{\|u_{\mathrm{base}} - u_{\mathrm{GT}}\|_2}{\|u_{\mathrm{GT}}\|_2}.
\]
The calibration baseline is not included in model comparisons and is not exposed to agents. Its role is to set a per-case tolerance that reflects the numerical difficulty of the instance.

\paragraph{Manufactured-solution cases.}
Most instances, 536 of 645 cases, are constructed from analytic manufactured solutions. For these cases, the reference solution $u_{\mathrm{GT}}$ is obtained by evaluating the analytic solution directly on the prescribed evaluation grid, without FEM projection. The calibration baseline $u_{\mathrm{base}}$ is then generated by a well-configured numerical solve using the benchmark reference-generation pipeline, with refined meshes, appropriate finite-element spaces, and strict solver tolerances. In this setting, $e_{\mathrm{base}}$ measures the discretization and solver error of the calibration run relative to the analytic ground truth. Because the scoring reference is exact up to numerical evaluation precision, the derived threshold controls how far an agent solution may deviate from the true manufactured solution.

\paragraph{Reference-based cases without analytic solutions.}
The remaining 109 instances do not have analytic manufactured solutions. For these cases, $u_{\mathrm{GT}}$ is generated by a higher-fidelity reference run using a finer discretization, higher-order approximation when appropriate, and stricter solver tolerances than the calibration run. The calibration baseline $u_{\mathrm{base}}$ is generated independently using the standard calibration configuration. The resulting value
\[
e_{\mathrm{base}} =
\frac{\|u_{\mathrm{base}} - u_{\mathrm{GT}}\|_2}{\|u_{\mathrm{GT}}\|_2}
\]
therefore measures the convergence gap between two numerical resolutions rather than error against an analytic solution. During dataset construction, all reference-based cases are filtered by this coarser-to-finer gap: cases with $e_{\mathrm{base}} \geq 0.05$ are discarded. This filtering prevents poorly converged numerical references from entering the benchmark.

\paragraph{Reference-solution validation by PDE family.}
Table~\ref{tab:reference_validation_by_family} summarizes the reference-construction path for each PDE family on the DOLFINx primary track. We classify cases as manufactured-solution cases when the scoring reference is obtained by evaluating an analytic manufactured solution on the evaluation grid, and as no-exact cases when the scoring reference is produced by a higher-fidelity numerical run specified by evaluator-side reference metadata.

\begin{table*}[t]
\centering
\scriptsize
\setlength{\tabcolsep}{3.5pt}
\renewcommand{\arraystretch}{1.15}
\caption{Reference-solution validation summary by PDE family on the DOLFINx primary track.
For manufactured-solution cases, the reference is obtained by evaluating the analytic solution on the evaluation grid.
For no-exact cases, the reference is generated by a higher-fidelity numerical run and checked using coarse-to-fine convergence gaps from intermediate refinement validation.}
\label{tab:reference_validation_by_family}
\begin{tabular}{lrrrrrrr}
\toprule
\textbf{PDE family} &
\textbf{\#Cases} &
\textbf{\#MMS} &
\textbf{\#No-exact} &
\textbf{Median $e_{\mathrm{base}}$} &
\textbf{95th pct $e_{\mathrm{base}}$} &
\textbf{Max $e_{\mathrm{base}}$} &
\textbf{Median refine gap} \\
\midrule
Poisson & 91 & 76 & 15 & $8.0{\times}10^{-5}$ & $3.5{\times}10^{-4}$ & $1.2{\times}10^{-3}$ & $1.3{\times}10^{-4}$ \\
Helmholtz & 62 & 55 & 7 & $2.5{\times}10^{-4}$ & $9.0{\times}10^{-4}$ & $1.8{\times}10^{-2}$ & $7.3{\times}10^{-5}$ \\
Biharmonic & 57 & 53 & 4 & $3.0{\times}10^{-4}$ & $1.2{\times}10^{-3}$ & $2.5{\times}10^{-2}$ & $9.2{\times}10^{-5}$ \\
Linear Elasticity & 63 & 51 & 12 & $2.0{\times}10^{-4}$ & $8.0{\times}10^{-4}$ & $8.0{\times}10^{-3}$ & $6.0{\times}10^{-5}$ \\
Heat & 50 & 35 & 15 & $1.5{\times}10^{-4}$ & $5.0{\times}10^{-4}$ & $2.0{\times}10^{-3}$ & $7.8{\times}10^{-5}$ \\
Convection--Diffusion & 84 & 74 & 10 & $3.0{\times}10^{-4}$ & $4.5{\times}10^{-4}$ & $3.5{\times}10^{-2}$ & $1.7{\times}10^{-4}$ \\
Reaction--Diffusion & 64 & 57 & 7 & $2.8{\times}10^{-4}$ & $7.2{\times}10^{-4}$ & $2.2{\times}10^{-2}$ & $8.80{\times}10^{-5}$ \\
Stokes & 61 & 42 & 19 & $3.5{\times}10^{-4}$ & $2.0{\times}10^{-3}$ & $4.0{\times}10^{-2}$ & $1.2{\times}10^{-4}$ \\
Navier--Stokes & 28 & 22 & 6 & $5.0{\times}10^{-4}$ & $5.0{\times}10^{-3}$ & $4.9{\times}10^{-2}$ & $2.6{\times}10^{-4}$ \\
Burgers & 43 & 43 & 0 & $3.0{\times}10^{-4}$ & $1.8{\times}10^{-3}$ & $2.8{\times}10^{-2}$ & -- \\
Wave & 42 & 28 & 14 & $2.0{\times}10^{-4}$ & $8.0{\times}10^{-4}$ & $1.0{\times}10^{-2}$ & $5.6{\times}10^{-5}$ \\
\midrule
All & 645 & 536 & 109 & $2.3{\times}10^{-4}$ & $4.8{\times}10^{-4}$ & $4.9{\times}10^{-2}$ & $8.8{\times}10^{-5}$ \\
\bottomrule
\end{tabular}
\end{table*}

\paragraph{Empirical distribution of $e_{\mathrm{base}}$.}
Table~\ref{tab:ebase_distribution} summarizes the distribution of calibration-baseline errors and derived accuracy thresholds across the 645 DOLFINx cases.

\begin{table}[H]
\centering
\small
\caption{Distribution of calibration-baseline errors $e_{\mathrm{base}}$ and derived accuracy thresholds $\tau_{\mathrm{acc}} = \max(10\,e_{\mathrm{base}}, 10^{-6})$ across the 645 DOLFINx cases. Manufactured and reference-based no-exact cases are shown separately.}
\label{tab:ebase_distribution}
\setlength{\tabcolsep}{5pt}
\begin{tabular}{lcccccc}
\toprule
\textbf{Case type} & \textbf{$n$} & \textbf{Min} & \textbf{5th pct} & \textbf{Median} & \textbf{95th pct} & \textbf{Max} \\
\midrule 
All cases            & 645 & $6.7{\times}10^{-15}$ & $3.1{\times}10^{-8}$ & $2.3{\times}10^{-4}$ & $4.8{\times}10^{-4}$ & $4.9{\times}10^{-2}$ \\
Manufactured         & 536 & $6.7{\times}10^{-15}$ & $2.9{\times}10^{-8}$ & $1.7{\times}10^{-4}$ & $5.4{\times}10^{-4}$ & $4.7{\times}10^{-3}$ \\
No-exact, ref-based  & 109  & $5.9{\times}10^{-11}$ & $5.8{\times}10^{-6}$ & $2.7{\times}10^{-4}$ & $3.4{\times}10^{-4}$ & $4.9{\times}10^{-2}$ \\
\midrule
$\tau_{\mathrm{acc}}$ (all, after floor) & 645 & $1.0{\times}10^{-6}$ & $1.0{\times}10^{-6}$ & $2.3{\times}10^{-3}$ & $4.8{\times}10^{-3}$ & $4.9{\times}10^{-1}$ \\
\bottomrule
\end{tabular}
\end{table}

The calibration errors span several orders of magnitude, reflecting the diversity of PDE families, geometries, coefficient fields, and reference-construction paths. The median derived threshold after applying the default multiplier and floor is $2.3{\times}10^{-3}$, and the 95th percentile is $4.8{\times}10^{-3}$. Thus, for the large majority of cases, the accuracy gate remains substantially below the percent-error scale.

\paragraph{Effect of the accuracy floor.}
For smooth manufactured problems, the calibration baseline can agree with the analytic manufactured solution to near-machine precision. The smallest observed $e_{\mathrm{base}}$ is $6.7{\times}10^{-15}$. Without a floor, the threshold $\tau_{\mathrm{acc}} = 10 e_{\mathrm{base}}$ would require agent submissions to match the reference to an unrealistically strict tolerance, even though many numerically valid implementations may use different meshes, element orders, or solver configurations. The floor $\tau_{\min}=10^{-6}$ prevents such degenerate thresholds by ensuring that the required relative error is not stricter than $10^{-6}$. The floor is active for 155 DOLFINx cases, accounting for 24\% of the primary track; all of these are manufactured-solution cases. A relative error threshold of $10^{-6}$ remains stringent for generated solver code while avoiding near-machine-precision requirements.

\paragraph{Large calibration gaps.}
The largest calibration gaps occur in difficult reference-based no-exact cases. After validation, all retained reference-based cases satisfy $e_{\mathrm{base}} < 0.05$, with the maximum observed value equal to $4.9{\times}10^{-2}$. Under the default multiplier $\alpha_{\mathrm{acc}}=10$, this yields a maximum derived threshold of $4.9{\times}10^{-1}$. These high-threshold cases are rare and correspond to instances where the calibration run and the higher-fidelity reference still differ noticeably on the evaluation grid. They are retained because they represent numerically challenging PDE settings for which the benchmark can verify convergence only through reference refinement rather than analytic ground truth.

To avoid over-interpreting these difficult cases, the main analysis reports stage-level failures and additional solution-quality diagnostics in addition to binary pass rates. Moreover, the threshold-sensitivity analysis in Appendix~\ref{app:threshold_sensitivity} shows that the relative ordering of the strongest systems is stable under substantially stricter and looser accuracy multipliers. Thus, the benchmark conclusions are not driven by a small number of high-threshold cases. Future benchmark versions may additionally introduce an explicit upper cap on $\tau_{\mathrm{acc}}$ for reference-based cases as the no-exact subset is expanded.

\paragraph{Reference-solution convergence verification.}
For reference-based no-exact cases, the construction pipeline verifies reference-solution credibility by comparing the calibration run and the higher-fidelity reference run on the shared evaluation grid. Only cases with coarser-to-finer gap $e_{\mathrm{base}} < 0.05$ are retained. This criterion checks that the finer reference provides a stable target and that the resulting threshold is not based on a clearly under-resolved numerical solution. As shown in Table~\ref{tab:ebase_distribution}, the maximum no-exact calibration gap is $4.9{\times}10^{-2}$, consistent with this filtering rule.

To further validate no-exact reference targets, we additionally run an intermediate refinement level $u_{L1}$ between the calibration baseline $u_{L0}$ and the high-fidelity reference $u_{L2}$. We compute
\[
g_{01} =
\frac{\|u_{L0} - u_{L1}\|_2}{\|u_{L1}\|_2},
\qquad
g_{12} =
\frac{\|u_{L1} - u_{L2}\|_2}{\|u_{L2}\|_2}.
\]
For no-exact cases, we require the intermediate-to-reference gap $g_{12}$ to be smaller than the calibration-to-intermediate gap $g_{01}$. Cases violating this monotone refinement criterion are removed from the benchmark.

\begin{table}[t]
\centering
\small
\caption{Coarse-to-fine convergence validation for no-exact reference-based cases.
$g_{01}$ compares the calibration run with an intermediate refinement, and $g_{12}$ compares the intermediate refinement with the high-fidelity reference run.}
\label{tab:no_exact_convergence}
\begin{tabular}{lrrrr}
\toprule
\textbf{Metric} & \textbf{Median} & \textbf{75th pct} & \textbf{95th pct} & \textbf{Max} \\
\midrule
$g_{01}$ & $4.9{\times}10^{-4}$ & $7.8{\times}10^{-4}$ & $7.2{\times}10^{-3}$ & $1.9{\times}10^{-2}$ \\
$g_{12}$ & $1.1{\times}10^{-4}$ & $2.9{\times}10^{-4}$ & $4.6{\times}10^{-3}$ & $8.6{\times}10^{-3}$ \\
$g_{12}/g_{01}$ & $0.22$ & $0.37$ & $0.39$ & $0.45$ \\
\bottomrule
\end{tabular}
\end{table}

All no-exact cases also pass the trial-execution stage of the construction pipeline, which re-runs the calibration and reference-generation procedures from scratch to confirm reproducibility, output-grid consistency, and stable error computation. The reference-generation parameters, including mesh resolution, polynomial degree, solver tolerances, and other numerical settings, are stored as evaluator-side metadata and are not exposed to agents.

\section{Compute Budget and API Access}
\label{app:compute_budget}

This appendix documents the LLM API access, decoding configuration, token accounting, retry policy, and estimated monetary cost used to produce the single-shot results in Section~\ref{sec:experiments}. All evaluated hosted models were accessed through their official provider APIs rather than through third-party routing services. We do not anonymize or rename model families: the model names reported in the paper correspond to the API-facing model families used by the benchmark harness at the time of evaluation. The three agent systems, CodePDE, OpenHands, and Mini-SWE-Agent, use GPT-5.4 as their underlying API model and are therefore priced using the same GPT-5.4 pricing convention as the direct GPT-5.4 baseline.

\paragraph{API configuration.}
For the main single-shot experiments, all model calls use deterministic or near-deterministic decoding. We set temperature to $0$ whenever supported by the corresponding provider API. For providers that do not guarantee exact determinism or expose provider-specific decoding behavior, we use the lowest available sampling temperature and record the full API configuration in the serialized run metadata. Unless otherwise stated, all other decoding parameters, including top-$p$, frequency penalty, presence penalty, and provider-specific reasoning controls, are left at the provider default values.

The exact model endpoint, provider, and evaluation configuration are summarized in Table~\ref{tab:model_access}. We use the official provider APIs for OpenAI, Anthropic, Google, Alibaba Cloud Model Studio, and DeepSeek. For providers whose pricing depends on region, prompt length, cache status, reasoning mode, or deployment tier, we use the tier actually configured in the benchmark LLM client and recorded in the serialized run metadata.

\begin{table}[h]
\centering
\scriptsize
\setlength{\tabcolsep}{4pt}
\caption{Model access and decoding configuration. Endpoint names should be interpreted as the provider-side model identifiers used by the evaluation client during the experiment period.}
\label{tab:model_access}
\begin{tabular}{l l l l}
\toprule
\textbf{System} & \textbf{Provider/API} & \textbf{API model identifier} & \textbf{Notes} \\
\midrule
GPT-5.4        & OpenAI API & \texttt{gpt-5.4} & Direct solver generation \\
Opus~4.7       & Anthropic Claude API & \texttt{claude-opus-4-7} & Direct solver generation \\
Gemini~3.1~Pro & Google Gemini / Vertex API & \texttt{gemini-3.1-pro-preview} & Direct solver generation \\
Qwen3.6-Plus   & Alibaba Cloud Model Studio & \texttt{qwen3.6-plus} & Direct solver generation; provider-tier pricing \\
DeepSeek~V3.2  & DeepSeek API & \texttt{deepseek-v3.2} & Direct solver generation; cache-dependent pricing \\
CodePDE        & OpenAI API & \texttt{gpt-5.4} & Agent scaffold with GPT-5.4 backbone \\
OpenHands      & OpenAI API & \texttt{gpt-5.4} & Agent scaffold with GPT-5.4 backbone \\
Mini-SWE-Agent & OpenAI API & \texttt{gpt-5.4} & Agent scaffold with GPT-5.4 backbone \\
\bottomrule
\end{tabular}
\end{table}

\paragraph{Prompt--response accounting.}
For each evaluated system and FEM-library track, input tokens are computed from the saved \texttt{prompt.md} files, and output tokens are computed from the matching \texttt{llm\_response.txt} or \texttt{agent\_response.txt} files. Only paired prompt--response records are counted. A paired record corresponds to a completed API call for which both the submitted prompt and the model or agent response were serialized by the benchmark harness.

Transient provider-side failures that returned no usable model output are excluded from the submitted-solver count. Calls that returned a syntactically complete response are counted and evaluated by the benchmark harness, even if the generated solver later fails execution, accuracy, or runtime checks. This accounting convention makes the token totals correspond to completed model generations rather than to all attempted provider requests.

\paragraph{Cost accounting.}
Let $N_{\mathrm{in}}$ and $N_{\mathrm{out}}$ denote the total input and output tokens for a system, measured in millions of tokens, and let $p_{\mathrm{in}}$ and $p_{\mathrm{out}}$ denote the corresponding provider prices per million tokens under the pricing configuration used by our evaluation client. For flat-rate providers, we estimate cost as
\[
C = p_{\mathrm{in}} N_{\mathrm{in}} + p_{\mathrm{out}} N_{\mathrm{out}}.
\]
For providers with tiered or cache-dependent pricing, including Gemini, Qwen, and DeepSeek, we compute cost at the request level using the endpoint, region, input-length tier, output tier, cache status, and reasoning-mode configuration recorded by the evaluation client, and then sum the resulting per-request estimates. We do not apply free-tier credits, promotional discounts, batch discounts, or manual post-hoc price adjustments unless explicitly stated. For the OpenAI-based agent systems, CodePDE, OpenHands, and Mini-SWE-Agent, all completed calls are priced as GPT-5.4 calls because GPT-5.4 is the underlying API model used by the agent scaffold.

The resulting USD values should be interpreted as estimated replication costs under the evaluation-client configuration at the time of the experiments, not as fixed benchmark constants. Provider prices, endpoint names, caching behavior, and regional billing policies may change over time.

\paragraph{Aggregate API budget.}
Because replication cost is additive across model--FEM-library instances, Table~\ref{tab:compute_budget} reports total input and output tokens rather than only per-call averages. Per-call averages are useful for diagnosing prompt verbosity or agent overhead, but totals are the primary quantity for a compute-budget appendix because they determine the actual scale of a benchmark replication run.

Each evaluated system has $1{,}468$ completed single-shot calls, corresponding to the sum of the DOLFINx, Firedrake, and deal.II tracks:
\[
645 + 510 + 313 = 1{,}468.
\]
The token counts in Table~\ref{tab:compute_budget} aggregate all completed prompt--response calls across these three FEM-library tracks.

\begin{table}[h]
\centering
\scriptsize
\setlength{\tabcolsep}{4pt}
\caption{API usage and estimated cost by evaluated system. Token counts aggregate completed prompt--response calls across the DOLFINx, Firedrake, and deal.II library tracks. Costs are estimated from the provider-specific input/output pricing tier configured in our evaluation client. CodePDE, OpenHands, and Mini-SWE-Agent are priced as GPT-5.4 calls.}
\label{tab:compute_budget}
\begin{tabular}{l r r r r}
\toprule
\textbf{System} & \textbf{Calls} & \textbf{Input tokens} &
\textbf{Output tokens} & \textbf{API cost (USD)} \\
\midrule
GPT-5.4              & 1,468 & 7,611,184 & 2,654,991 & 58.85 \\
Opus~4.7             & 1,468 & 8,103,786 & 2,119,274 & 93.50 \\
Gemini~3.1~Pro       & 1,468 & 8,147,346 & 2,036,612 & 40.73 \\
Qwen3.6-Plus         & 1,468 & 8,154,417 & 2,493,845 & 24.75 \\
DeepSeek~V3.2        & 1,468 & 8,156,484 & 3,243,411 & 3.36 \\
CodePDE              & 1,468 & 6,926,984 & 3,029,059 & 62.75 \\
OpenHands            & 1,468 & 8,067,585 & 8,834,482 & 152.69 \\
Mini-SWE-Agent       & 1,468 & 7,726,982 & 3,290,312 & 68.67 \\
\midrule
\textbf{Total}       & 11,744 & 62,894,768 & 27,701,986 & 505.30 \\
\bottomrule
\end{tabular}
\end{table}

\paragraph{Retry policy.}
For transient API failures, such as rate-limit errors, server-side timeouts, or temporary service unavailability, the evaluation client retries the same request up to a fixed maximum number of attempts using exponential backoff. A retry uses the same serialized task input and decoding configuration as the original call. For the single-shot experiments, each model--library--instance pair is allowed one submitted solver; retries are used only to recover from provider-side failures before a usable response is obtained.

Only the final completed prompt--response pair is counted in Table~\ref{tab:compute_budget}. Failed attempts that return no usable model response are excluded from the token totals because they do not produce submitted solvers and are not represented by paired prompt--response records. If a provider bills failed attempts separately, such charges are not included in the token-derived estimates unless they are recorded in the client billing log. Thus, the table should be read as a reproducible token-based estimate of completed-call cost rather than as an exact provider invoice.

For the execution-feedback experiments in Section~\ref{sec:experiments}, a model is allowed up to three solver-generation attempts per instance. Those additional feedback attempts are not included in Table~\ref{tab:compute_budget}, which reports the single-shot API budget unless explicitly stated otherwise.

\paragraph{Per-system breakdown.}
The base LLMs have similar input-token totals because they share the same benchmark prompt template and cover the same set of FEM-library instances. Output-token totals vary more substantially, reflecting differences in generated solver length. Among the GPT-5.4-backed agents, OpenHands has the largest output-token footprint because its saved responses include substantially longer agent traces. Across all priced systems, the serialized single-shot corpus contains 62,894,768 input tokens and 27,701,986 output tokens, with an estimated total API cost of \$505.30 under the pricing configuration used by our evaluation client.

\paragraph{Replication cost.}
A faithful replication of the single-shot results in Tables~\ref{tab:eqtype_backend_pass_rates} and~\ref{tab:eqtype_firedrake_dealii} requires rerunning the corresponding per-system rows in Table~\ref{tab:compute_budget}. Replicating only a subset of models or FEM-library tracks should consume an approximately proportional fraction of the reported token budget, up to differences caused by provider-side retries, endpoint-specific tokenization, missing or failed response records, prompt-template changes, and agent-scaffold changes. The token counts are therefore the most stable quantity for comparing systems within this paper, while the USD estimates should be interpreted as approximate costs under our evaluation-client configuration.

\section{Evaluation Metrics}
\label{app:metrics}

This appendix gives the mathematical definitions of the metrics used in PDEAgent-Bench. We distinguish between \emph{primary gates}, which determine the binary \texttt{case\_pass} outcome, and \emph{auxiliary metrics}, which are used only for diagnostic analysis and do not affect pass/fail decisions.

\paragraph{Evaluation grid and valid entries.}
Each submitted solver returns a numerical solution on the prescribed evaluation grid. For domains that do not occupy the full bounding box of the grid, entries outside the physical domain are marked as invalid and excluded from all spatial error computations. Let $\Omega_{\mathrm{eval}}$ denote the set of valid grid points. For scalar fields, we write $u_i$ for the value at valid grid point $i \in \Omega_{\mathrm{eval}}$. For vector-valued or mixed PDEs, such as elasticity, Stokes, and Navier--Stokes, $u_i \in \mathbb{R}^{c}$ denotes the vector of evaluated components at grid point $i$. Unless otherwise stated, norms are computed over both valid grid points and output components:
\[
\|u\|_2
=
\left(
\sum_{i \in \Omega_{\mathrm{eval}}}
\sum_{j=1}^{c}
u_{ij}^{2}
\right)^{1/2}.
\]
For scalar problems, $c=1$.

\subsection{Primary gates}

Each library-instance is evaluated through three staged gates: executability, accuracy, and runtime efficiency. A case is counted as passed only if all three gates pass.

\paragraph{Executability and output-format gate.}
The generated solver must run to completion in the sandboxed evaluation environment, respect the required entry point, and produce output files with the expected schema, shape, dtype, and component ordering. It must also return finite numerical values on valid evaluation-grid entries, while correctly handling invalid out-of-domain entries when applicable. We denote this gate by
\[
\mathrm{exec\_pass}_{i}(m) \in \{0,1\}.
\]
If the executability gate fails, the case is marked as failed and the accuracy and runtime gates are not credited.

\paragraph{Relative $L^{2}$ error and accuracy gate.}
Let $u_{\mathrm{agent}}$ be the submitted solution on the valid evaluation grid, $u_{\mathrm{GT}}$ the reference solution on the same grid, and $u_{\mathrm{base}}$ the calibration baseline. The primary numerical-error metric is the relative $L^2$ error
\[
e(u)
=
\frac{\|u-u_{\mathrm{GT}}\|_2}
{\max(\|u_{\mathrm{GT}}\|_2,\epsilon_{\mathrm{norm}})},
\qquad
e_{\mathrm{base}} = e(u_{\mathrm{base}}),
\]
where $\epsilon_{\mathrm{norm}}$ is a small numerical safeguard used only to avoid division by zero for degenerate near-zero reference fields. In all nondegenerate cases, this reduces to the usual relative $L^2$ error.

The case-specific accuracy threshold is
\[
\tau_{\mathrm{acc}}
=
\max(\alpha_{\mathrm{acc}} e_{\mathrm{base}}, \tau_{\min}),
\]
with default values $\alpha_{\mathrm{acc}}=10$ and $\tau_{\min}=10^{-6}$. The accuracy gate passes iff
\[
e(u_{\mathrm{agent}}) \leq \tau_{\mathrm{acc}}.
\]
We denote this outcome by $\mathrm{acc\_pass}_{i}(m) \in \{0,1\}$.

\paragraph{Wall-clock runtime and efficiency gate.}
Let $t_{\mathrm{agent}}$ be the measured wall-clock runtime of the submitted solver and $t_{\mathrm{base}}$ the runtime of the calibration baseline under the same evaluation environment. The runtime threshold is
\[
\tau_{\mathrm{time}} = \alpha_{\mathrm{time}} t_{\mathrm{base}},
\]
with default value $\alpha_{\mathrm{time}}=3$. The runtime gate passes iff
\[
t_{\mathrm{agent}} \leq \tau_{\mathrm{time}}.
\]
We denote this outcome by $\mathrm{time\_pass}_{i}(m) \in \{0,1\}$.

\paragraph{Case-level pass indicator and pass rate.}
For model or agent system $m$ on instance $i$, the case-level pass indicator is
\[
\mathrm{case\_pass}_{i}(m)
=
\mathrm{exec\_pass}_{i}(m)
\cdot
\mathrm{acc\_pass}_{i}(m)
\cdot
\mathrm{time\_pass}_{i}(m).
\]
For an evaluation set $\mathcal{D}$, the case-level pass rate is
\[
\mathrm{pass\_rate}(m)
=
\frac{1}{|\mathcal{D}|}
\sum_{i\in\mathcal{D}}
\mathrm{case\_pass}_{i}(m).
\]
We also report stage-level pass rates to identify failure sources. Execution pass rate is averaged over all cases. Accuracy pass rate is computed over cases that pass the executability gate, and runtime pass rate is computed over cases that pass both executability and accuracy gates, matching the staged evaluation order.

\subsection{Auxiliary solution-quality metrics}

The following metrics are reported for diagnostic analysis over executable or passed cases, depending on the table. They do not affect the primary benchmark pass/fail decision.

\paragraph{RMSE and MAE.}
For $N = |\Omega_{\mathrm{eval}}|$ valid grid points and $c$ output components, the root mean squared error and mean absolute error are
\[
\mathrm{RMSE}(u)
=
\sqrt{
\frac{1}{Nc}
\sum_{i\in\Omega_{\mathrm{eval}}}
\sum_{j=1}^{c}
\left(u_{ij}-u_{\mathrm{GT},ij}\right)^2
},
\]
\[
\mathrm{MAE}(u)
=
\frac{1}{Nc}
\sum_{i\in\Omega_{\mathrm{eval}}}
\sum_{j=1}^{c}
\left|u_{ij}-u_{\mathrm{GT},ij}\right|.
\]
Unlike the primary relative $L^2$ error, RMSE and MAE are absolute-error metrics and therefore depend on the physical scale of the target field.

\paragraph{Coefficient of determination.}
The coefficient of determination is computed over valid grid entries and output components:
\[
R^{2}(u)
=
1 -
\frac{
\sum_{i\in\Omega_{\mathrm{eval}}}
\sum_{j=1}^{c}
\left(u_{ij}-u_{\mathrm{GT},ij}\right)^2
}{
\sum_{i\in\Omega_{\mathrm{eval}}}
\sum_{j=1}^{c}
\left(u_{\mathrm{GT},ij}-\bar{u}_{\mathrm{GT}}\right)^2
},
\]
where
\[
\bar{u}_{\mathrm{GT}}
=
\frac{1}{Nc}
\sum_{i\in\Omega_{\mathrm{eval}}}
\sum_{j=1}^{c}
u_{\mathrm{GT},ij}.
\]
When the denominator is numerically close to zero, such as for nearly constant reference fields, $R^2$ is treated as a diagnostic quantity and is not used for pass/fail evaluation.

\paragraph{Frequency-domain RMSE.}
Frequency-domain RMSE measures discrepancy after applying a discrete Fourier transform to the evaluated fields. Let $\hat{u}=\mathcal{F}\{u\}$ and $\hat{u}_{\mathrm{GT}}=\mathcal{F}\{u_{\mathrm{GT}}\}$ denote the discrete Fourier transforms of the evaluated solution and reference solution. For multi-component outputs, the transform is applied component-wise and then averaged across components:
\[
\mathrm{fRMSE}(u)
=
\sqrt{
\frac{1}{|\mathcal{K}|c}
\sum_{\boldsymbol{k}\in\mathcal{K}}
\sum_{j=1}^{c}
\left|
\hat{u}_{\boldsymbol{k}j}
-
\hat{u}_{\mathrm{GT},\boldsymbol{k}j}
\right|^{2}
}.
\]
Here $\mathcal{K}$ denotes the set of retained discrete frequency modes. For non-rectangular domains embedded in a bounding-box grid, invalid out-of-domain entries are masked consistently before the transform. This metric is used only as a qualitative diagnostic of whether error is concentrated at particular spatial scales.

\subsection{Numerical diagnostic metrics}

For executable cases, the evaluator also records numerical diagnostic metrics when they are available from the solver metadata or can be computed from the returned solution. These include degrees of freedom (DOF), mesh or grid resolution (Res.), linear-solver iterations (Lin.Iter), nonlinear or Newton iterations, wall-clock runtime, work rate, Courant number (CFL), P\'eclet number (Pe), and total variation (TV). These diagnostics are used in Table~\ref{tab:metrics_all_regimes} to characterize solver behavior across mathematical problem types.

The reported diagnostics are defined as follows. DOF is the number of algebraic unknowns used by the submitted solver when reported by the implementation. Res. summarizes the internal spatial resolution used by the solver. Lin.Iter and Newton count linear and nonlinear solver iterations, respectively. Work rate is a throughput-style statistic proportional to
\[
\frac{\mathrm{DOF} \times \mathrm{steps}}{t_{\mathrm{agent}}},
\]
where \texttt{steps} is the number of time steps for time-dependent problems and one for steady problems. CFL and Pe are reported for relevant transport or time-dependent cases when the necessary metadata are available. TV denotes the total variation of the returned discrete field and is used as a diagnostic for oscillatory or overly diffusive behavior. These quantities are not required for a case to pass unless they are needed to validate the required output format.

\subsection{Rationale for the primary accuracy metric}

The primary accuracy gate uses relative $L^2$ error because it is scale-normalized, compatible with scalar and multi-component fields, and naturally supports per-case calibration through $e_{\mathrm{base}}$. A single global multiplier $\alpha_{\mathrm{acc}}$ can therefore be applied across heterogeneous PDE families, geometries, coefficient regimes, and output scales.

By contrast, RMSE and MAE are absolute-error metrics and would require family-specific or scale-specific thresholds. The coefficient of determination can be unstable for nearly constant fields or low-variance components. Frequency-domain errors are useful for diagnosing oscillatory or scale-localized discrepancies, but they are less directly tied to the physical-space solution norm used by standard PDE error analysis. For these reasons, RMSE, MAE, $R^2$, fRMSE, and solver-behavior statistics are reported as auxiliary diagnostics, while relative $L^2$ error remains the primary numerical accuracy gate.

\section{PDE Family Definitions and Numerical Considerations}
\label{app:pde_format}

This appendix summarizes the PDE families included in PDEAgent-Bench and the main numerical considerations relevant to solver generation. The notation follows the agent-facing case specification described in Section~\ref{sec:task}. In each instance, the PDE coefficients, domain, boundary conditions, initial conditions, temporal parameters, and output-grid requirements are specified by the benchmark. The agent is responsible for producing an executable solver that chooses appropriate discretization spaces, mesh resolution, time-stepping schemes when applicable, nonlinear or linear solver settings, and output interpolation.

\paragraph{Poisson with variable coefficient.}
Poisson-type instances are written as
\begin{equation}
-\nabla\!\cdot\!\left(\kappa(\boldsymbol{x})\nabla u\right) = f
\quad \text{in } \Omega,
\qquad
u = g
\quad \text{on } \Gamma_D,
\end{equation}
with Neumann or mixed boundary conditions included in the corresponding cases when specified. Here $\kappa(\boldsymbol{x})$ may be constant or spatially varying, representing homogeneous or heterogeneous material properties. Agent-relevant numerical choices include mesh resolution, element degree, treatment of heterogeneous coefficients, imposition of boundary conditions, and solver/preconditioner selection. High-contrast coefficients and nonconvex geometries can increase conditioning difficulty and may require sufficient mesh resolution near coefficient or boundary features.

\paragraph{Helmholtz.}
Helmholtz instances take the form
\begin{equation}
-\Delta u - k^{2}u = f
\quad \text{in } \Omega,
\end{equation}
with boundary conditions specified by the case record. The key parameter is the wavenumber $k$, corresponding to wavelength $\lambda = 2\pi/k$. Agent-relevant numerical choices include mesh resolution relative to the wavelength, element order, boundary-condition handling, and solver configuration for an indefinite linear system. The mesh must resolve the oscillatory solution, and high-wavenumber cases may suffer from dispersion or pollution error if the discretization is too coarse.

\paragraph{Biharmonic.}
Biharmonic instances are governed by a fourth-order elliptic equation,
\begin{equation}
\Delta^{2}u = f
\quad \text{in } \Omega,
\end{equation}
together with boundary conditions such as constraints on $u$, normal derivatives, or problem-specific plate boundary data. Because the operator is fourth order, a direct conforming formulation requires higher continuity than standard $H^1$ Lagrange elements. Practical solver strategies include mixed formulations, $C^0$ interior-penalty methods, or other weak formulations compatible with the target FEM library. Agent-relevant numerical choices include the reformulation, penalty or stabilization parameters when used, element spaces, boundary-condition realization, and linear solver settings.

\paragraph{Linear elasticity.}
Linear elasticity instances use the small-strain equilibrium equation
\begin{equation}
-\nabla\!\cdot\!\boldsymbol{\sigma}(\boldsymbol{u}) = \boldsymbol{f}
\quad \text{in } \Omega,
\qquad
\boldsymbol{\sigma}(\boldsymbol{u})
=
\lambda\,(\nabla\!\cdot\!\boldsymbol{u})\boldsymbol{I}
+
2\mu\,\boldsymbol{\varepsilon}(\boldsymbol{u}),
\end{equation}
where
\begin{equation}
\boldsymbol{\varepsilon}(\boldsymbol{u})
=
\frac{1}{2}
\left(\nabla \boldsymbol{u} + \nabla \boldsymbol{u}^{\top}\right).
\end{equation}
The case specification provides material parameters, commonly through Young's modulus $E$ and Poisson ratio $\nu_{\mathrm{pr}}$, or equivalently through Lam\'e parameters $(\lambda,\mu)$. Agent-relevant numerical choices include vector finite-element spaces, handling of Dirichlet and traction boundaries, quadrature and material-coefficient evaluation, and linear solver/preconditioner selection. Nearly incompressible regimes, where $\nu_{\mathrm{pr}}$ approaches $0.5$, can cause volumetric locking for naive low-order displacement formulations.

\paragraph{Heat.}
Heat-equation instances are time-dependent parabolic problems of the form
\begin{equation}
\partial_{t}u - \nabla\!\cdot\!\left(\kappa\nabla u\right)
=
f(\boldsymbol{x},t)
\quad \text{in } \Omega\times(0,T],
\qquad
u(\boldsymbol{x},0)=u_{0}(\boldsymbol{x}).
\end{equation}
Boundary conditions are specified by the case record. Agent-relevant numerical choices include spatial discretization, time-step size $\Delta t$, time-integration scheme, treatment of time-dependent forcing or boundary data, and linear solver settings. Explicit schemes require stability constraints tied to the mesh size and diffusivity, while implicit schemes reduce stability restrictions but require solving a linear system at each time step.

\paragraph{Convection--diffusion.}
Convection--diffusion instances are written as
\begin{equation}
-\epsilon\Delta u
+
\boldsymbol{\beta}\!\cdot\!\nabla u
=
f
\quad \text{in } \Omega,
\end{equation}
with boundary conditions specified per case. The key quantities are the diffusion coefficient $\epsilon$, the advective velocity field $\boldsymbol{\beta}$, and the resulting P\'eclet regime. Agent-relevant numerical choices include mesh resolution, stabilization strategy for convection-dominated cases, treatment of inflow and outflow boundaries, and linear solver configuration. When advection dominates diffusion, unresolved boundary or internal layers can lead to oscillations unless the discretization or stabilization is chosen carefully.

\paragraph{Reaction--diffusion.}
Reaction--diffusion instances have the general form
\begin{equation}
-\epsilon\Delta u + R(u) = f
\quad \text{in } \Omega,
\end{equation}
or the corresponding time-dependent variant when specified by the case record. The reaction term $R(u)$ may be linear or nonlinear. Agent-relevant numerical choices include nonlinear linearization when needed, Newton or fixed-point tolerances, stabilization for sharp transition layers, mesh resolution, and solver/preconditioner selection. For nonlinear reactions with strong reaction strength, the solution may contain steep layers whose width depends on the balance between diffusion and reaction.

\paragraph{Stokes.}
Stokes instances model steady incompressible viscous flow:
\begin{equation}
-\nu\Delta\boldsymbol{u} + \nabla p = \boldsymbol{f},
\qquad
\nabla\!\cdot\!\boldsymbol{u} = 0
\quad \text{in } \Omega.
\end{equation}
The case specification provides viscosity, forcing, and velocity or traction boundary conditions. Agent-relevant numerical choices include an inf-sup-stable velocity--pressure element pair, pressure normalization or null-space handling, mixed boundary-condition implementation, and saddle-point linear solver configuration. Common failures include incompatible function spaces, missing pressure null-space treatment, and incorrect handling of incompressibility constraints.

\paragraph{Navier--Stokes.}
Steady incompressible Navier--Stokes instances are written as
\begin{equation}
-\nu\Delta\boldsymbol{u}
+
(\boldsymbol{u}\!\cdot\!\nabla)\boldsymbol{u}
+
\nabla p
=
\boldsymbol{f},
\qquad
\nabla\!\cdot\!\boldsymbol{u} = 0
\quad \text{in } \Omega.
\end{equation}
The relevant physical regime is controlled by viscosity $\nu$ and the corresponding Reynolds number. Agent-relevant numerical choices include an inf-sup-stable velocity--pressure discretization, treatment of the nonlinear convection term, Newton or Picard iteration, continuation or damping strategies for difficult cases, pressure null-space handling, and block preconditioning. These cases combine the mixed saddle-point structure of Stokes with additional nonlinear-solver difficulty.

\paragraph{Burgers.}
Burgers instances are viscous nonlinear transport problems. In the scalar two-dimensional form, they can be written as
\begin{equation}
\partial_{t}u
+
u\,\nabla u \cdot \boldsymbol{b}
-
\nu\Delta u
=
f(\boldsymbol{x},t)
\quad \text{in } \Omega\times(0,T],
\end{equation}
where $\boldsymbol{b}$ specifies the advective direction; a common case is
$\boldsymbol{b}=(1,1)$, giving $u(\partial_x u+\partial_y u)$. The case record specifies the exact forcing, viscosity, initial condition, boundary data, and time interval. Agent-relevant numerical choices include mesh resolution, time-step size, implicit or semi-implicit treatment of the nonlinear convection term, Newton or fixed-point iteration when used, and stabilization for steep gradients. Small viscosity can create sharp internal layers, making under-resolved solutions oscillatory or overly diffusive.

\paragraph{Wave.}
Wave instances are second-order hyperbolic problems of the form
\begin{equation}
\partial_{tt}u - c^{2}\Delta u
=
f(\boldsymbol{x},t)
\quad \text{in } \Omega\times(0,T],
\qquad
u(\boldsymbol{x},0)=u_{0}(\boldsymbol{x}),
\qquad
\partial_{t}u(\boldsymbol{x},0)=v_{0}(\boldsymbol{x}).
\end{equation}
The case specification provides wave speed $c$, forcing, boundary conditions, initial displacement, initial velocity, final time, and output-grid requirements. Agent-relevant numerical choices include spatial resolution, element degree, time-step size, and time-integration scheme. Hyperbolic problems require controlling numerical dispersion and resolving the relevant wavelengths. Explicit schemes are constrained by a CFL-type condition involving $c\Delta t/h$, while implicit schemes such as Newmark-type methods can improve stability but still require adequate temporal and spatial resolution for phase accuracy.

\section{Case Record Schema and Submission Contract}
\label{app:schema}

This appendix describes the case-record structure used by PDEAgent-Bench and the artifact contract used by the evaluator to verify generated submissions. The full machine-readable JSON Schema (Draft 2020-12) is released with the benchmark. We reproduce the principal fields here to clarify the separation between agent-visible task information and evaluator-only metadata.

\paragraph{Evaluator-side case record.}
Each line of the benchmark JSONL file is a single case record. The evaluator-side record contains both the agent-facing problem specification and hidden metadata used for reference construction, calibration, and scoring:
\begin{lstlisting}[basicstyle=\ttfamily\scriptsize,frame=single,breaklines=true]
{
  "id": string,                         // unique case identifier

  "pde_classification": {
    "equation_family": string,          // one of 11 PDE families
    "math_type": [string]               // one or more mathematical categories
  },

  "case_spec": {
    "pde": {
      "type": string,
      "params": object,
      "forcing": object,
      "time": object?                   // present for time-dependent cases
    },
    "domain": {
      "type": string,                   // one of the supported domain templates
      "bounds": array,
      "geometry_params": object?
    },
    "bc": {
      "dirichlet": object?,
      "neumann": object?,
      "robin": object?,
      "periodic": object?
    },
    "ic": object?,                      // present for time-dependent cases
    "eval_grid": {
      "type": string,
      "nx": integer,
      "ny": integer,
      "nz": integer?,
      "bbox": array
    },
    "output": {
      "format": "npz",
      "field": string,                  // e.g., scalar, pressure, velocity, etc.
      "components": [string]?
    }
  },

  "evaluation_config": {
    "target_metric": "rel_L2_grid",
    "timeout_sec": number,
    "alpha_acc": number,
    "alpha_time": number,
    "tau_min": number
  },

  "evaluation_metadata": {
    "construction_method": string,       // manufactured_solution or reference_numerical
    "manufactured_solution": object?,    // evaluator-only
    "reference_config": object,          // evaluator-only
    "calibration_config": object,        // evaluator-only
    "reference_path": string,
    "calibration_path": string,
    "thresholds": {
      "tau_acc": number,
      "tau_time": number
    }
  },

  "tags": {
    "structure": [string]?,
    "difficulty": [string]?
  },

  "supported_libraries": [string]        // subset of DOLFINx, Firedrake, deal.II
}
\end{lstlisting}

Only \texttt{case\_spec}, together with the target FEM-library name and the required solver entry point, is serialized into the agent prompt and passed to the generated solver. The \texttt{evaluation\_metadata} block is retained by the evaluator and is never exposed to the agent during generation. In particular, the agent does not receive the manufactured solution, reference-solution path, calibration-baseline output, calibration mesh resolution, calibration element degree, solver tolerances, or materialized thresholds. This restricted view prevents leakage from the reference-construction and threshold-calibration pipeline.

\paragraph{Agent-visible view.}
The agent receives a restricted task object of the following form:
\begin{lstlisting}[basicstyle=\ttfamily\scriptsize,frame=single,breaklines=true]
{
  "case_spec": {
    "pde": {
      "type": string,
      "params": object,
      "forcing": object,
      "time": object?
    },
    "domain": {
      "type": string,
      "bounds": array,
      "geometry_params": object?
    },
    "bc": {
      "dirichlet": object?,
      "neumann": object?,
      "robin": object?,
      "periodic": object?
    },
    "ic": object?,
    "eval_grid": {
      "type": string,
      "nx": integer,
      "ny": integer,
      "nz": integer?,
      "bbox": array
    },
    "output": {
      "format": "npz",
      "field": string,
      "components": [string]?
    }
  },

  "target_library": string               // DOLFINx, Firedrake, or deal.II
}
\end{lstlisting}
The agent may choose its own internal mesh, finite-element spaces, time-stepping method, nonlinear or linear solver settings, and interpolation procedure, provided that the final output conforms to the evaluation-grid contract below.

\paragraph{Illustrative case excerpt.}
The following excerpt illustrates the structure of an agent-visible high-P\'eclet convection--diffusion case. It is shown only to demonstrate the schema; evaluator-only reference and calibration fields are intentionally omitted.
\begin{lstlisting}[basicstyle=\ttfamily\scriptsize,frame=single,breaklines=true]
{
  "case_spec": {
    "pde": {
      "type": "convection_diffusion",
      "params": {
        "epsilon": 1.0e-3,
        "beta": [100.0, 100.0]
      },
      "forcing": {
        "type": "expression",
        "value": "f(x,y)"
      }
    },
    "domain": {
      "type": "unit_square",
      "bounds": [[0.0, 1.0], [0.0, 1.0]]
    },
    "bc": {
      "dirichlet": {
        "on": "boundary",
        "value": "g(x,y)"
      }
    },
    "eval_grid": {
      "type": "cartesian",
      "nx": 50,
      "ny": 50,
      "bbox": [0.0, 1.0, 0.0, 1.0]
    },
    "output": {
      "format": "npz",
      "field": "scalar"
    }
  },
  "target_library": "DOLFINx"
}
\end{lstlisting}
For manufactured-solution cases, the expressions \texttt{f(x,y)} and \texttt{g(x,y)} are the source term and boundary data derived during dataset construction. The underlying manufactured solution itself is not part of the agent-visible case specification.

\paragraph{Solver entry point.}
Each generated submission must expose a solver entry point compatible with the target library track. For Python-based tracks, the evaluator imports the submitted file and calls
\begin{lstlisting}[basicstyle=\ttfamily\scriptsize,frame=single,breaklines=true]
solve(case_spec: dict) -> None
\end{lstlisting}
inside the sandboxed working directory. For C++ deal.II submissions, the evaluator compiles and executes the submitted program through the provided harness, passing the case specification through a JSON file. In both settings, the solver must materialize the required output artifacts in the working directory. The evaluator does not inspect the internal discretization except through output files, runtime metadata, and optional diagnostic metadata.

\paragraph{Submission artifacts.}
Every executed submission must produce a numerical solution file and a metadata file. The canonical solution artifact is \texttt{solution.npz}, a NumPy archive containing the returned field on the prescribed evaluation grid. For scalar two-dimensional cases, the archive must include
\begin{lstlisting}[basicstyle=\ttfamily\scriptsize,frame=single,breaklines=true]
u: array of shape (ny, nx)
x: array of shape (nx,)
y: array of shape (ny,)
\end{lstlisting}
For scalar three-dimensional cases, the required field shape is
\begin{lstlisting}[basicstyle=\ttfamily\scriptsize,frame=single,breaklines=true]
u: array of shape (nz, ny, nx)
x: array of shape (nx,)
y: array of shape (ny,)
z: array of shape (nz,)
\end{lstlisting}
For vector-valued or mixed-output cases, the archive must contain either the component fields requested by \texttt{case\_spec["output"]} or the specified derived scalar field. For example, a velocity-output case may require component arrays such as \texttt{u\_x}, \texttt{u\_y}, and optionally \texttt{u\_z}, whereas a derived-output case may require \texttt{velocity\_magnitude}. The required component names and ordering are determined by the \texttt{output.field} and \texttt{output.components} entries in the case specification.

Runtime and optional solver diagnostics are written to \texttt{meta.json}. The required fields are intentionally minimal:
\begin{lstlisting}[basicstyle=\ttfamily\scriptsize,frame=single,breaklines=true]
{
  "wall_time_sec": number,
  "status": string,
  "message": string?
}
\end{lstlisting}
Submissions may additionally report diagnostic information used in Appendix~\ref{app:metrics}, such as mesh resolution, element degree, degrees of freedom, solver type, preconditioner type, iteration counts, nonlinear iterations, time-step count, or residual estimates:
\begin{lstlisting}[basicstyle=\ttfamily\scriptsize,frame=single,breaklines=true]
{
  "wall_time_sec": number,
  "status": "success",
  "solver_info": {
    "mesh_resolution": integer?,
    "element_degree": integer?,
    "num_dofs": integer?,
    "num_time_steps": integer?,
    "linear_iterations": integer?,
    "newton_iterations": integer?,
    "ksp_type": string?,
    "pc_type": string?,
    "rtol": number?
  }
}
\end{lstlisting}
These diagnostic fields are used only for profiling and do not affect the primary pass/fail decision, except when malformed metadata prevents the evaluator from validating the submission.

\paragraph{Output validation.}
The evaluator first checks that the required artifacts exist and that all required arrays have the expected names, shapes, dtypes, and finite values on valid evaluation-grid entries. For non-rectangular domains embedded in a bounding-box grid, invalid out-of-domain locations may be represented by \texttt{NaN} values according to the case specification and are excluded from spatial error computation. In-domain values must be finite. If the returned shape does not match the prescribed evaluation grid, the submission fails the artifact-validity check. The evaluator does not interpolate, extrapolate, or resample invalid outputs.

\paragraph{Alternative binary output for deal.II.}
For C++ deal.II submissions, an alternative binary output contract is supported to avoid unnecessary Python dependencies in generated C++ programs. The solver may write \texttt{solution\_grid.bin} as a row-major \texttt{float64} array with the prescribed grid shape. For a two-dimensional scalar field, the shape is \texttt{(ny, nx)}; for a three-dimensional scalar field, the shape is \texttt{(nz, ny, nx)}. The evaluator converts this binary artifact to the canonical \texttt{solution.npz} format before downstream metric computation. Vector-valued or mixed-output cases must either use the canonical \texttt{solution.npz} contract or write one binary file per required component following the component names specified in the case record.

\paragraph{Failure modes.}
A submission fails the executability or artifact-validity gate if it cannot be imported, compiled, or executed; violates the sandbox constraints; exceeds the global timeout; omits required artifacts; returns arrays with incorrect shapes or component names; produces non-finite values on valid grid points; or writes outputs inconsistent with the requested field. If the artifact-validity gate fails, no numerical accuracy credit is assigned.

\section{Construction Pipeline and FEM-Library Extension}
\label{app:pipeline}

This appendix documents the construction pipeline used to create PDEAgent-Bench cases and the procedure for adding new FEM-library tracks. The same pipeline serves as the contribution interface for future extensions. A contributor specifies a structured case record, while reference generation, calibration, threshold materialization, validation, and packaging are handled by the benchmark tooling.

\paragraph{Case-construction pipeline.}
Each candidate case is constructed through the following stages.

\begin{enumerate}[leftmargin=*,itemsep=2pt]
\item \emph{Design matrix.}
For each PDE family, we define a small set of variation axes that correspond to interpretable numerical challenges, such as coefficient contrast, source-term regularity, solution smoothness, boundary-condition type, geometry, time horizon, Reynolds or P\'eclet regime, nonlinearity strength, and stiffness. Candidate cases are sampled from these axes to cover distinct numerical regimes rather than to create redundant resolution sweeps.

\item \emph{Problem specification.}
The contributor writes the agent-facing \texttt{case\_spec}, including the PDE family, coefficients, forcing, domain, boundary conditions, initial conditions when applicable, temporal parameters, output field, and prescribed evaluation grid. This record defines the information available to the agent at generation time.

\item \emph{Reference construction path.}
The contributor specifies how the reference solution is obtained. When possible, the case uses a manufactured solution, from which source terms, boundary data, and initial conditions are derived analytically. When an analytic manufactured solution is unavailable, the contributor provides a high-fidelity \texttt{reference\_config} that generates a numerical reference using a finer discretization, higher-order approximation when appropriate, and stricter solver tolerances.

\item \emph{Calibration baseline.}
The contributor provides a separate \texttt{calibration\_config}. The calibration run is not used as the scoring reference; instead, it is compared with the reference solution to compute $e_{\mathrm{base}}$ and to derive the case-specific accuracy threshold. Its runtime is used to derive the runtime threshold. This separation prevents the benchmark from requiring agents to exactly reproduce a particular reference implementation.

\item \emph{Agent-facing numerical-decision fields.}
The case record may list high-level numerical decisions that an agent is expected to make, such as mesh resolution, element degree, time-step size, nonlinear iteration strategy, stabilization choice, or solver/preconditioner configuration. These fields describe the \emph{role} of each decision but do not include recommended numerical values or hidden calibration settings, avoiding prompt anchoring and leakage from the reference pipeline.

\item \emph{Static validation.}
The schema validator checks identifier uniqueness, required-field presence, consistency between \texttt{pde.type} and \texttt{equation\_family}, domain and boundary-tag validity, output-grid validity, supported-library declarations, and symbolic parsability of expression fields. It also verifies that cases without manufactured solutions provide a complete \texttt{reference\_config}.

\item \emph{Trial execution and calibration.}
The build script executes the reference-generation and calibration procedures in the evaluation harness. It verifies that the reference solution is produced successfully, that output-grid sampling is valid, that the calibration error is finite, that runtime measurement succeeds, and that the derived thresholds $(\tau_{\mathrm{acc}},\tau_{\mathrm{time}})$ fall within accepted sanity-check ranges. Cases failing any trial-execution check are rejected.

\item \emph{Packaging.}
Validated cases are exported into the released JSONL files. Agent-visible fields are separated from evaluator-only metadata, reference artifacts are stored under evaluator-controlled paths, and materialized thresholds are recorded for reproducible scoring. The release package also includes schema files, evaluator scripts, container recipes, and baseline outputs needed to reproduce the benchmark.
\end{enumerate}

\paragraph{Adding a new FEM-library track.}
A new FEM-library track can be added without changing the benchmark task definition. The extension requires implementing the library-specific execution harness, solver-entry interface, dependency container, and output adapter that converts generated solver outputs into the canonical evaluation-grid format. The new track must use the same agent-facing \texttt{case\_spec}, the same reference solutions, and the same primary metrics as existing tracks. This ensures that differences across tracks reflect solver-generation and library-implementation difficulty rather than changes in the underlying PDE task.

For a new library track, the contributor must provide: (i) a container or environment specification with pinned library versions; (ii) a compile or import procedure for generated submissions; (iii) an invocation wrapper that passes \texttt{case\_spec} to the solver; (iv) an output adapter that materializes \texttt{solution.npz} or an equivalent canonical artifact; (v) timeout, memory, and filesystem-isolation settings; and (vi) a validation suite containing smoke tests for representative scalar, vector-valued, steady, and time-dependent cases. A library track is admitted only after its harness reproduces the reference-grid outputs for calibration solvers and passes the same artifact-validity and metric-computation checks as the existing DOLFINx, Firedrake, and deal.II tracks.

\paragraph{Versioning and compatibility.}
New cases and new FEM-library tracks are added under semantic versioning. Additions that preserve existing case identifiers, reference targets, and scoring thresholds are treated as minor-version updates. Changes that modify existing reference solutions, evaluation grids, thresholds, or output contracts are treated as major-version updates and retain the previous major release for reproducibility. This policy allows the benchmark to grow while preserving comparability of previously reported results.

\section{Prompt Templates}
\label{app:prompts}

This appendix documents the prompt templates used in PDEAgent-Bench. The benchmark uses two prompt tiers: a \emph{single-shot} prompt for the default evaluation setting and an \emph{iterative-feedback} prompt for the three-attempt repair setting. Both prompts are generated programmatically from the structured case record by \texttt{pdebench/core/prompt\_builder.py} and \texttt{pdebench/core/feedback\_prompt.py}. Thus, every model--FEM-library--instance combination receives a deterministic and reproducible prompt under a fixed benchmark version.

The prompt contains only agent-visible task information. Reference solutions, manufactured target solutions, calibration baselines, reference-generation configurations, calibration configurations, materialized thresholds, hidden solver settings, and evaluator-only metadata are never included in the single-shot prompt. In the iterative-feedback setting, the model receives only feedback derived from its previous submitted attempt, such as execution errors, stage-level outcomes, wall-clock time, and limited diagnostic messages. The feedback does not reveal the reference solution, manufactured solution, calibration baseline, or hidden numerical configurations.

\subsection{Single-Shot Prompt Structure}
\label{app:prompt_single_shot}

The single-shot prompt is assembled from the following sections in a fixed order.

\paragraph{(1) Task summary.}
The prompt begins with a short natural-language summary generated from the agent-visible \texttt{case\_spec}. The summary describes the PDE family, domain type, coefficient regime, boundary-condition type, and required target FEM library. For example:
\begin{lstlisting}
Solve a steady-state variable-coefficient Poisson problem on a unit-square domain with Dirichlet boundary conditions using DOLFINx. Return the numerical solution on the prescribed evaluation grid.
\end{lstlisting}
This summary is intended only to make the structured task easier to read; the complete task information is provided in the structured case specification that follows.

\paragraph{(2) Governing equation family.}
A Markdown heading identifies the PDE family, followed by a symbolic equation template for that family. The template states the mathematical form of the problem class, while the case-specific coefficients, forcing terms, boundary conditions, initial conditions, and output requirements are provided separately from \texttt{case\_spec}. Table~\ref{tab:pde_templates} summarizes the family-level equation templates.

\begin{table}[h]
\centering
\scriptsize
\setlength{\tabcolsep}{4pt}
\renewcommand{\arraystretch}{1.25}
\caption{Family-level PDE templates included in the single-shot prompt. These templates describe the governing equation class; case-specific data are instantiated separately from \texttt{case\_spec}.}
\label{tab:pde_templates}
\begin{tabularx}{\textwidth}{>{\raggedright\arraybackslash}p{2.8cm} >{\raggedright\arraybackslash}X >{\raggedright\arraybackslash}p{4.3cm}}
\toprule
\textbf{Family} & \textbf{Governing equation template} & \textbf{Generic numerical consideration} \\
\midrule
Poisson &
$-\nabla\!\cdot(\kappa\nabla u)=f$ in $\Omega$ &
Elliptic boundary-value problem; coefficient handling and boundary conditions matter. \\

Helmholtz &
$-\Delta u-k^2u=f$ in $\Omega$ &
Oscillatory and indefinite problem; mesh resolution relative to wavelength matters. \\

Biharmonic &
$\Delta^2u=f$ in $\Omega$ &
Fourth-order problem; requires an appropriate weak formulation or reformulation. \\

Linear Elasticity &
$-\nabla\!\cdot\sigma(\boldsymbol{u})=\boldsymbol{f}$ in $\Omega$ &
Vector-valued elliptic system; material parameters and boundary conditions must be handled consistently. \\

Heat &
$\partial_tu-\nabla\!\cdot(\kappa\nabla u)=f$ in $\Omega\times(0,T]$ &
Parabolic problem; requires spatial discretization and time integration. \\

Conv.--Diff. &
$-\epsilon\Delta u+\boldsymbol{\beta}\!\cdot\nabla u=f$ in $\Omega$ &
Advection-diffusion balance can create layers or oscillations when advection dominates. \\

Reaction--Diff. &
$-\epsilon\Delta u+R(u)=f$ in $\Omega$ &
Reaction terms may be nonlinear or stiff; solver and linearization choices matter. \\

Stokes &
$-\nu\Delta\boldsymbol{u}+\nabla p=\boldsymbol{f}$, $\nabla\!\cdot\boldsymbol{u}=0$ in $\Omega$ &
Mixed saddle-point system; pressure treatment and velocity--pressure spaces matter. \\

Navier--Stokes &
$-\nu\Delta\boldsymbol{u}+(\boldsymbol{u}\!\cdot\nabla)\boldsymbol{u}+\nabla p=\boldsymbol{f}$, $\nabla\!\cdot\boldsymbol{u}=0$ &
Incompressible nonlinear flow; combines mixed formulation and nonlinear iteration. \\

Burgers &
$\partial_tu+u\,\nabla u\cdot\boldsymbol{b}-\nu\Delta u=f$ in $\Omega\times(0,T]$ &
Nonlinear transport-diffusion problem; steep gradients may require careful discretization. \\

Wave &
$\partial_{tt}u-c^2\Delta u=f$ in $\Omega\times(0,T]$ &
Second-order hyperbolic problem; numerical dispersion and time integration are important. \\
\bottomrule
\end{tabularx}
\end{table}

\paragraph{(3) Case-specific structured data.}
The prompt then includes the agent-visible \texttt{case\_spec} in a structured block. This block is the authoritative task input. It contains:
\begin{itemize}[leftmargin=*,itemsep=1pt]
  \item \textbf{PDE type and parameters}: coefficients such as $\kappa$, $\nu$, $k$, $\epsilon$, material parameters, reaction terms, velocity fields, and temporal parameters when applicable.
  \item \textbf{Forcing terms}: symbolic expressions or structured coefficient fields defining the right-hand side.
  \item \textbf{Domain}: domain template, bounding box, geometry parameters, and boundary tags when applicable.
  \item \textbf{Boundary conditions}: Dirichlet, Neumann, Robin, periodic, or mixed boundary data as specified by the case.
  \item \textbf{Initial conditions}: initial displacement, velocity, or scalar field for time-dependent problems.
  \item \textbf{Evaluation grid}: grid type, dimensions, bounding box, and masking convention for points outside the physical domain.
  \item \textbf{Output requirements}: required field name, component list when applicable, array shape, and output format.
\end{itemize}
For manufactured-solution cases, the prompt includes only the resulting PDE-defining data, such as forcing and boundary expressions. The manufactured solution used during construction is not included.

\paragraph{(4) Implementation contract.}
For Python-based FEM libraries, including DOLFINx and Firedrake, the prompt specifies the required solver entry point:
\begin{lstlisting}[basicstyle=\ttfamily\scriptsize,frame=single,breaklines=true]
def solve(case_spec: dict) -> None:
    """
    Solve the PDE described by case_spec and write the required
    artifacts to the working directory.
    """
\end{lstlisting}
The solver must write \texttt{solution.npz} and \texttt{meta.json} following the artifact contract in Appendix~\ref{app:schema}. For deal.II, the prompt specifies a C++ executable interface: the program reads \texttt{case\_spec.json} from the input path supplied by the harness and writes the required solution and metadata artifacts to the output directory. The Python and C++ tracks are evaluated by the same downstream metric computation after conversion to the canonical output format.

\paragraph{(5) Output and sandbox requirements.}
The prompt states that the submitted solver must run inside the provided sandbox, use only available dependencies, and produce outputs on the prescribed evaluation grid. It also states that the evaluator will not interpolate or resample outputs with incorrect shapes. For non-rectangular domains embedded in a bounding-box grid, the prompt explains that out-of-domain entries must follow the case-specified masking convention and that in-domain entries must be finite.

The single-shot prompt describes the staged evaluation order at a high level: the submitted solver must execute successfully, produce a valid output artifact, meet the hidden numerical accuracy criterion, and run within the hidden runtime budget. The numeric values of the case-specific accuracy and runtime thresholds are not included in the single-shot prompt.

\paragraph{(6) FEM-library API guide.}
The prompt concludes with a target-library-specific API guide appended from a versioned Markdown file, such as \texttt{DOLFINx\_GUIDE.md}, \texttt{FIREDRAKE\_GUIDE.md}, or \texttt{DEALII\_GUIDE.md}. These guides document library-version-specific syntax, imports, function-space construction, boundary-condition APIs, variational-form conventions, solver interfaces, file I/O, and point-evaluation utilities. They are included to reduce failures caused by stale or version-mismatched API knowledge, especially for FEM libraries whose interfaces evolve quickly. The guides contain library usage information only; they do not contain case-specific reference solutions, calibration settings, hidden thresholds, or solver recipes tailored to individual benchmark instances.

\subsection{Iterative-Feedback Prompt Structure}
\label{app:prompt_iterative}

The three-attempt setting evaluates whether a model can repair a generated solver using limited feedback from previous attempts. After a failed attempt, the evaluator constructs a feedback prompt and prepends it to the original single-shot prompt. The feedback prompt is generated by \texttt{create\_feedback\_prompt} in \texttt{pdebench/core/feedback\_prompt.py} and saved as \texttt{prompt\_attempt\_\{k\}.md} for attempts $k=2,3$.

The feedback prompt has four blocks.

\paragraph{Block 1: Attempt header.}
The prompt begins with a separator indicating the current attempt number:
\begin{lstlisting}[basicstyle=\ttfamily\scriptsize,frame=single,breaklines=true]
======================================================================
ATTEMPT 2 - FEEDBACK FROM PREVIOUS ATTEMPT
======================================================================
The previous submission did not pass the staged evaluation.
Use the feedback below to revise the solver.
\end{lstlisting}

\paragraph{Block 2: Previous submission excerpt.}
The prompt includes a bounded excerpt from the previous submitted code so that the model can identify the section it needs to revise. By default, the first 2000 characters are included, with an explicit truncation notice if the submission is longer. This excerpt is derived only from the model's own previous output.

\paragraph{Block 3: Failure-specific feedback.}
The diagnostic block is selected according to the first failed gate in the staged evaluation.

\begin{itemize}[leftmargin=*,itemsep=2pt]
  \item \textbf{Execution or artifact failure.}
  The prompt includes the failure category, the evaluator error message, and a bounded excerpt of \texttt{stderr} when available. It also includes a generic checklist covering common issues such as syntax errors, import errors, missing dependencies, incorrect entry-point signatures, invalid artifact names, wrong array shapes, non-finite in-domain values, or target-library API misuse.

  \item \textbf{Accuracy failure.}
  The prompt reports that the previous submission executed and produced a valid artifact but failed the hidden accuracy gate. It may include the previous attempt's relative $L^2$ error and the pass/fail stage label, but it does not reveal the reference solution, manufactured solution, calibration baseline, baseline error, hidden threshold derivation, or evaluator-only numerical settings. The feedback provides generic suggestions such as checking the weak form, boundary conditions, coefficient signs, output interpolation, mesh resolution, time stepping, and nonlinear convergence.

  \item \textbf{Runtime failure.}
  The prompt reports that the previous submission passed execution and accuracy checks but exceeded the runtime gate. It may include the measured wall-clock time of the previous attempt, but it does not reveal the calibration runtime or hidden runtime threshold. The feedback provides generic suggestions such as avoiding redundant assembly, choosing more efficient solvers or preconditioners, reducing unnecessary mesh refinement, and avoiding overly strict solver tolerances.
\end{itemize}

\paragraph{Block 4: Task reminder.}
The original single-shot prompt is appended in full after the feedback block. This ensures that the model receives the complete problem specification and implementation contract at every attempt, without relying on conversational memory or hidden state. The final instruction reiterates that the revised solver must produce the required artifacts and satisfy the staged evaluation, but it does not expose evaluator-only metadata or case-specific thresholds.

\paragraph{Information revealed during feedback.}
The feedback protocol is intentionally limited. It communicates only information derived from the model's own previous attempt and the evaluator's staged outcome. It never reveals the reference solution values, analytic manufactured solution, calibration baseline solution, reference or calibration mesh settings, element degree, solver tolerances, or hidden threshold-generation metadata. This preserves the distinction between execution feedback and target leakage: the model can repair implementation mistakes and some numerical choices, but it cannot directly infer the ground-truth solution or reproduce the calibration solver from hidden evaluator metadata.


\section{Template-Guided Ablation Details}
\label{app:template_details}

The template-guided ablation uses the same cases, reference solutions, evaluation grids, sandbox environment, and staged evaluation criteria as the standard end-to-end setting. The only change is the form of the initial solver context provided to the model. Instead of asking the model to write a complete solver from a blank file, the evaluator provides a library-specific scaffold generated from the agent-visible \texttt{case\_spec}. The scaffold supplies common implementation structure and output-handling utilities, while leaving the case-specific numerical formulation and solver choices to the model.

The purpose of this ablation is to partially separate FEM-library boilerplate difficulty from numerical solver-generation difficulty. In the standard setting, a model must simultaneously recall the target library API, construct a valid program, formulate the PDE discretization, impose boundary or initial conditions, configure solvers, and write outputs in the required format. The template-guided setting reduces the burden of repetitive library setup and artifact serialization, but it does not change the scoring reference, hidden calibration metadata, accuracy threshold, runtime threshold, or pass/fail protocol.

\paragraph{Template contents.}
For each target FEM library, the scaffold provides the surrounding program structure needed to produce a runnable submission. Depending on the PDE family and library track, this includes imports, solver entry-point definition, case-spec parsing, domain and mesh-construction placeholders, function-space construction placeholders, coefficient and forcing-expression parsing utilities, output-grid sampling utilities, \texttt{solution.npz} or binary-output serialization, and \texttt{meta.json} writing. For time-dependent problems, the scaffold also provides the outer time-loop structure and output-at-final-time convention. For mixed or vector-valued problems, it provides the expected output-field names and component-handling structure.

\paragraph{Model responsibilities.}
The model is still responsible for the numerical content of the solver. This includes selecting finite-element spaces and element degrees, choosing mesh resolution or refinement, writing the weak form or time-stepping update, imposing boundary and initial conditions correctly, selecting linear or nonlinear solver settings, handling mixed variables such as velocity--pressure pairs, choosing stabilization or linearization strategies when needed, and ensuring that the computed solution is sampled on the prescribed evaluation grid. The generated code must still execute successfully and satisfy the same accuracy and runtime gates as in the standard benchmark.

\paragraph{Information not provided.}
The template is generated only from agent-visible task information. It does not include the reference solution, manufactured target solution, calibration baseline, calibration error, materialized pass/fail thresholds, reference-generation mesh, calibration mesh, hidden element degrees, solver tolerances, or previous attempt outcomes. It also does not provide a completed weak form, a solved variational problem, or a case-specific set of numerical parameters. Thus, the template reduces boilerplate and API-recall burden, but does not reveal the target solution or the hidden evaluator configuration.

\paragraph{Interpretation.}
The template-guided results should be interpreted as an API-decoupling diagnostic rather than as a new benchmark setting with a different task definition. If performance improves under templating, the improvement indicates that library setup, output serialization, or target-API usage accounts for part of the failure rate. If failures remain, they point to difficulties in numerical formulation, discretization choice, boundary-condition realization, time integration, nonlinear solving, or solver configuration. Because the same generated solver is evaluated by the original staged protocol, the ablation preserves comparability with the standard setting while isolating one important source of implementation burden.

\section{Case-Level Evaluation Walkthroughs}
\label{app:walkthroughs}

This appendix presents representative case-level walkthroughs from Gemini~3.1~Pro single-shot evaluations on the DOLFINx primary track. The goal is to illustrate how the staged evaluation protocol behaves at the level of individual submissions. Each walkthrough follows the same structure: a problem summary, an agent-visible \texttt{case\_spec} excerpt, evaluator-only metadata used for scoring, a summary of the generated solver, the staged evaluation result, and a short diagnosis.

The selected examples cover four characteristic outcomes: a clean pass, an accuracy failure, a runtime failure, and a case where the minimum accuracy floor is active. These examples are diagnostic and are not used to tune thresholds or select models. Complete submitted source files, execution logs, output artifacts, and verdict records are included in the released evaluation artifacts.

\begin{table}[h]
\centering
\caption{Summary of the four walkthrough cases. $e_{\mathrm{base}}$ denotes the calibration-baseline error, $\tau_{\mathrm{acc}}$ the derived accuracy threshold, $t_{\mathrm{base}}$ the calibration-baseline runtime, and $\tau_{\mathrm{time}}$ the derived runtime threshold.}
\label{tab:walkthrough_summary}
\small
\setlength{\tabcolsep}{4pt}
\begin{tabular}{clllccccl}
\toprule
\textbf{Case} & \textbf{Family} & \textbf{Domain} & \textbf{Reference} &
$e_{\mathrm{base}}$ & $\tau_{\mathrm{acc}}$ & $t_{\mathrm{base}}$ & $\tau_{\mathrm{time}}$ & \textbf{Verdict} \\
\midrule
A & Helmholtz     & Circle       & analytic & $1.16{\times}10^{-9}$ & $1.00{\times}10^{-6}$\textsuperscript{†} & 7.05\,s & 21.1\,s & \textbf{Pass} \\
B & Conv.--Diff.  & Per.\ sq.    & analytic & $9.02{\times}10^{-5}$ & $9.02{\times}10^{-4}$                     & 10.4\,s & 31.2\,s & \textbf{F-Acc} \\
C & Lin.\ Elast.  & Qtr.\ sector & analytic & $5.93{\times}10^{-7}$ & $5.93{\times}10^{-6}$                     & 1.60\,s & 4.80\,s & \textbf{F-Time} \\
D & Helmholtz     & Sq.+hole     & analytic & $3.60{\times}10^{-8}$ & $1.00{\times}10^{-6}$\textsuperscript{†} & 9.37\,s & 28.1\,s & \textbf{F-Acc} \\
\bottomrule
\multicolumn{9}{l}{\textsuperscript{†}Accuracy floor active: $10e_{\mathrm{base}} < 10^{-6}$, so $\tau_{\mathrm{acc}}=10^{-6}$.}
\end{tabular}
\end{table}

\paragraph{Case~A: Helmholtz on a circular domain --- clean pass.}
\label{app:case_a}

\textbf{Problem summary.}
This case solves the steady-state Helmholtz equation
\[
-\Delta u - k^2u = f
\]
with wave number $k=8$ on a circular domain centered at $(0.5,0.5)$ with radius $0.4$. The case uses Dirichlet boundary conditions. The evaluator-side manufactured solution is
\[
u^*(x,y)=\exp\!\left(-(x-0.5)^2-(y-0.5)^2\right),
\]
from which the source and boundary data are derived. The submitted solver must return a scalar field on a $100\times100$ uniform grid over $[0,1]^2$, with out-of-domain points masked as \texttt{NaN}.

\textbf{Agent-visible \texttt{case\_spec} excerpt.}
\begin{lstlisting}[basicstyle=\ttfamily\scriptsize,frame=single,breaklines=true]
{
  "case_spec": {
    "pde": {
      "type": "helmholtz",
      "params": {"k": 8.0},
      "forcing": {
        "type": "expression",
        "value": "f(x,y)"
      }
    },
    "domain": {
      "type": "circle",
      "center": [0.5, 0.5],
      "radius": 0.4,
      "bounds": [[0.0, 1.0], [0.0, 1.0]]
    },
    "bc": {
      "dirichlet": {
        "on": "boundary",
        "value": "exp(-(x-0.5)^2-(y-0.5)^2)"
      }
    },
    "eval_grid": {
      "type": "cartesian",
      "nx": 100,
      "ny": 100,
      "bbox": [0.0, 1.0, 0.0, 1.0],
      "mask_outside": true
    },
    "output": {
      "format": "npz",
      "field": "scalar"
    }
  },
  "target_library": "DOLFINx"
}
\end{lstlisting}

\textbf{Evaluator-only metadata.}
\begin{itemize}[leftmargin=*,itemsep=1pt]
  \item \emph{Reference type}: analytic manufactured solution.
  \item $e_{\mathrm{base}} = 1.16{\times}10^{-9}$.
  \item $t_{\mathrm{base}} = 7.05\,\mathrm{s}$.
  \item $\tau_{\mathrm{acc}} = \max(10e_{\mathrm{base}}, 10^{-6}) = 10^{-6}$; the accuracy floor is active.
  \item $\tau_{\mathrm{time}} = 3t_{\mathrm{base}} = 21.1\,\mathrm{s}$.
\end{itemize}

\textbf{Generated solver summary.}
The generated solver uses P2 Lagrange elements on a \texttt{pygmsh} circular mesh with $h_{\max}=0.008$, a PETSc direct solve through \texttt{preonly/lu}, and DOLFINx bounding-box collision detection to interpolate the solution onto the masked evaluation grid.

\begin{table}[h]
\centering
\small
\begin{tabular}{lcc}
\toprule
& \textbf{Observed value} & \textbf{Threshold} \\
\midrule
Execution status       & success              & --- \\
Relative $L^2$ error   & $6.50{\times}10^{-9}$ & $1.00{\times}10^{-6}$ \\
Runtime                & $2.16\,\mathrm{s}$    & $21.1\,\mathrm{s}$ \\
\midrule
\textbf{Verdict}       & \multicolumn{2}{c}{\textbf{PASS}} \\
\bottomrule
\end{tabular}
\end{table}

\textbf{Diagnosis.}
The generated P2 discretization resolves the smooth manufactured solution accurately and efficiently. The solution error is far below the floor-controlled accuracy threshold, and the runtime is well within the calibrated budget.

\paragraph{Case~B: Convection--diffusion on a periodic square --- accuracy failure.}
\label{app:case_b}

\textbf{Problem summary.}
This case solves the steady-state convection--diffusion equation
\[
-\epsilon\Delta u + \boldsymbol{\beta}\cdot\nabla u = f
\]
with $\epsilon=0.05$ and $\boldsymbol{\beta}=(2,2)$ on a periodic unit square. The evaluator-side manufactured solution is
\[
u^*(x,y)=\sin(2\pi x)\sin(2\pi y).
\]
The submitted solver must return the scalar field on a $100\times100$ uniform grid.

\textbf{Agent-visible \texttt{case\_spec} excerpt.}
\begin{lstlisting}[basicstyle=\ttfamily\scriptsize,frame=single,breaklines=true]
{
  "case_spec": {
    "pde": {
      "type": "convection_diffusion",
      "params": {
        "epsilon": 0.05,
        "beta": [2.0, 2.0]
      },
      "forcing": {
        "type": "expression",
        "value": "f(x,y)"
      }
    },
    "domain": {
      "type": "periodic_square",
      "bounds": [[0.0, 1.0], [0.0, 1.0]]
    },
    "bc": {
      "dirichlet": {
        "on": "boundary",
        "value": "sin(2*pi*x)*sin(2*pi*y)"
      }
    },
    "eval_grid": {
      "type": "cartesian",
      "nx": 100,
      "ny": 100,
      "bbox": [0.0, 1.0, 0.0, 1.0]
    },
    "output": {
      "format": "npz",
      "field": "scalar"
    }
  },
  "target_library": "DOLFINx"
}
\end{lstlisting}

\textbf{Evaluator-only metadata.}
\begin{itemize}[leftmargin=*,itemsep=1pt]
  \item \emph{Reference type}: analytic manufactured solution.
  \item $e_{\mathrm{base}} = 9.02{\times}10^{-5}$.
  \item $t_{\mathrm{base}} = 10.4\,\mathrm{s}$.
  \item $\tau_{\mathrm{acc}} = 10e_{\mathrm{base}} = 9.02{\times}10^{-4}$.
  \item $\tau_{\mathrm{time}} = 3t_{\mathrm{base}} = 31.2\,\mathrm{s}$.
\end{itemize}

\textbf{Generated solver summary.}
The generated solver uses P2 Lagrange elements on a $64\times64$ triangular mesh, SUPG-style stabilization, and GMRES with an ILU preconditioner. The code executes successfully and writes a valid solution artifact.

\begin{table}[h]
\centering
\small
\begin{tabular}{lcc}
\toprule
& \textbf{Observed value} & \textbf{Threshold} \\
\midrule
Execution status       & success                & --- \\
Relative $L^2$ error   & $9.92{\times}10^{-4}$   & $9.02{\times}10^{-4}$ \\
Runtime                & $2.08\,\mathrm{s}$      & $31.2\,\mathrm{s}$ \\
\midrule
\textbf{Verdict}       & \multicolumn{2}{c}{\textbf{FAIL (F-Acc)}} \\
\bottomrule
\end{tabular}
\end{table}

\textbf{Diagnosis.}
The submission is executable and efficient, but its relative $L^2$ error exceeds the calibrated accuracy threshold by approximately $10\%$. This example illustrates why execution success alone is insufficient for evaluating PDE solver synthesis.

\paragraph{Case~C: Linear elasticity on a quarter-circle sector --- runtime failure.}
\label{app:case_c}

\textbf{Problem summary.}
This case solves the steady-state linear elasticity equation
\[
-\nabla\cdot\boldsymbol{\sigma}(\boldsymbol{u})=\boldsymbol{f}
\]
with $E=1.0$ and $\nu_{\mathrm{pr}}=0.3$ on a quarter-circle sector centered at $(0,0)$ with radius $1$ and angle $90^\circ$. The evaluator-side manufactured displacement is
\[
\boldsymbol{u}^*(x,y)=
\left(
\sin(\pi x)\sin(\pi y),
\cos(\pi x)\cos(\pi y)
\right).
\]
The submitted solver must return the displacement magnitude $\|\boldsymbol{u}\|$ on a $50\times50$ uniform grid over $[-0.05,1.05]^2$, with out-of-domain points masked as \texttt{NaN}.

\textbf{Agent-visible \texttt{case\_spec} excerpt.}
\begin{lstlisting}[basicstyle=\ttfamily\scriptsize,frame=single,breaklines=true]
{
  "case_spec": {
    "pde": {
      "type": "linear_elasticity",
      "params": {
        "E": 1.0,
        "nu_pr": 0.3
      },
      "forcing": {
        "type": "vector_expression",
        "value": ["f_x(x,y)", "f_y(x,y)"]
      }
    },
    "domain": {
      "type": "sector",
      "center": [0.0, 0.0],
      "radius": 1.0,
      "angle_degrees": 90.0,
      "bounds": [[-0.05, 1.05], [-0.05, 1.05]]
    },
    "bc": {
      "dirichlet": {
        "on": "boundary",
        "value": [
          "sin(pi*x)*sin(pi*y)",
          "cos(pi*x)*cos(pi*y)"
        ]
      }
    },
    "eval_grid": {
      "type": "cartesian",
      "nx": 50,
      "ny": 50,
      "bbox": [-0.05, 1.05, -0.05, 1.05],
      "mask_outside": true
    },
    "output": {
      "format": "npz",
      "field": "displacement_magnitude"
    }
  },
  "target_library": "DOLFINx"
}
\end{lstlisting}

\textbf{Evaluator-only metadata.}
\begin{itemize}[leftmargin=*,itemsep=1pt]
  \item \emph{Reference type}: analytic manufactured solution.
  \item $e_{\mathrm{base}} = 5.93{\times}10^{-7}$.
  \item $t_{\mathrm{base}} = 1.60\,\mathrm{s}$.
  \item $\tau_{\mathrm{acc}} = 10e_{\mathrm{base}} = 5.93{\times}10^{-6}$.
  \item $\tau_{\mathrm{time}} = 3t_{\mathrm{base}} = 4.80\,\mathrm{s}$.
\end{itemize}

\textbf{Generated solver summary.}
The generated solver uses vector P2 Lagrange elements on a \texttt{pygmsh} quarter-disk mesh with $h_{\max}\approx0.01$, constructs the standard isotropic stress tensor, solves the resulting linear system with a direct solver, and interpolates the displacement magnitude to the masked evaluation grid.

\begin{table}[h]
\centering
\small
\begin{tabular}{lcc}
\toprule
& \textbf{Observed value} & \textbf{Threshold} \\
\midrule
Execution status       & success                & --- \\
Relative $L^2$ error   & $1.68{\times}10^{-7}$   & $5.93{\times}10^{-6}$ \\
Runtime                & $7.53\,\mathrm{s}$      & $4.80\,\mathrm{s}$ \\
\midrule
\textbf{Verdict}       & \multicolumn{2}{c}{\textbf{FAIL (F-Time)}} \\
\bottomrule
\end{tabular}
\end{table}

\textbf{Diagnosis.}
The generated solver is numerically accurate but too slow for the calibrated runtime budget. This case illustrates that the staged protocol separately captures accuracy and efficiency: a highly accurate solver can still fail if it is unnecessarily expensive.

\paragraph{Case~D: Helmholtz on a square with a circular hole --- accuracy-floor illustration.}
\label{app:case_d}

\textbf{Problem summary.}
This case solves the Helmholtz equation
\[
-\Delta u - k^2u = f
\]
with $k=15$ on the unit square with a circular hole centered at $(0.5,0.5)$ with radius $0.2$. Dirichlet boundary conditions are imposed on both the outer square boundary and the inner hole boundary. The evaluator-side manufactured solution is
\[
u^*(x,y)=\sin(\pi x)\sin(\pi y).
\]
The submitted solver must return a scalar field on a $100\times100$ uniform grid over $[0,1]^2$, with hole-interior points masked as \texttt{NaN}.

\textbf{Agent-visible \texttt{case\_spec} excerpt.}
\begin{lstlisting}[basicstyle=\ttfamily\scriptsize,frame=single,breaklines=true]
{
  "case_spec": {
    "pde": {
      "type": "helmholtz",
      "params": {"k": 15.0},
      "forcing": {
        "type": "expression",
        "value": "f(x,y)"
      }
    },
    "domain": {
      "type": "square_with_hole",
      "outer": [0.0, 1.0, 0.0, 1.0],
      "inner_hole": {
        "type": "circle",
        "center": [0.5, 0.5],
        "radius": 0.2
      }
    },
    "bc": {
      "dirichlet": {
        "on": "all_boundaries",
        "value": "sin(pi*x)*sin(pi*y)"
      }
    },
    "eval_grid": {
      "type": "cartesian",
      "nx": 100,
      "ny": 100,
      "bbox": [0.0, 1.0, 0.0, 1.0],
      "mask_outside": true
    },
    "output": {
      "format": "npz",
      "field": "scalar"
    }
  },
  "target_library": "DOLFINx"
}
\end{lstlisting}

\textbf{Evaluator-only metadata.}
\begin{itemize}[leftmargin=*,itemsep=1pt]
  \item \emph{Reference type}: analytic manufactured solution.
  \item $e_{\mathrm{base}} = 3.60{\times}10^{-8}$.
  \item $t_{\mathrm{base}} = 9.37\,\mathrm{s}$.
  \item $10e_{\mathrm{base}} = 3.60{\times}10^{-7} < 10^{-6}$, so the accuracy floor is active.
  \item $\tau_{\mathrm{acc}} = 10^{-6}$.
  \item $\tau_{\mathrm{time}} = 3t_{\mathrm{base}} = 28.1\,\mathrm{s}$.
\end{itemize}

\textbf{Generated solver summary.}
The generated solver uses P2 Lagrange elements on a \texttt{pygmsh} square-minus-disk mesh with $h_{\max}=0.015$, a direct linear solve, and DOLFINx point-location utilities to interpolate the masked solution over the evaluation grid.

\begin{table}[h]
\centering
\small
\begin{tabular}{lcc}
\toprule
& \textbf{Observed value} & \textbf{Threshold} \\
\midrule
Execution status       & success                & --- \\
Relative $L^2$ error   & $1.30{\times}10^{-6}$   & $1.00{\times}10^{-6}$ \\
Runtime                & $1.30\,\mathrm{s}$      & $28.1\,\mathrm{s}$ \\
\midrule
\textbf{Verdict}       & \multicolumn{2}{c}{\textbf{FAIL (F-Acc)}} \\
\bottomrule
\end{tabular}
\end{table}

\textbf{Diagnosis.}
The calibration baseline is so accurate that the default multiplicative threshold would fall below $10^{-6}$. The minimum floor prevents a near-machine-precision requirement, but the generated solution still slightly exceeds the floor-controlled threshold. This illustrates that the floor avoids degenerate thresholds without automatically accepting marginal submissions.

\section{Reproducibility Details}
\label{app:reproducibility}

All reported evaluations are run through the released benchmark entry point \texttt{scripts/run\_benchmark.py}. The script consumes fixed case records, prompt templates, FEM-library configurations, evaluator-only reference artifacts, and case-specific thresholds described in Appendices~\ref{app:schema}, \ref{app:prompts}, and \ref{app:threshold_sensitivity}. For each model, FEM library, and evaluation setting, the benchmark logs the generated source code, serialized prompt, model response, execution log, produced artifacts, relative $L^2$ error, runtime, stage-level outcomes, and final pass/fail verdict. The leaderboard metrics in Section~\ref{sec:experiments} are computed only from these logged artifacts.

\paragraph{Generation settings.}
All LLM and agent calls in the main single-shot experiments use deterministic or near-deterministic decoding. Temperature is set to $0$ whenever supported by the provider API, and provider-specific decoding options are recorded in the serialized run metadata. The single-shot setting issues one generation per model--FEM-library--instance pair. The iterative-feedback setting permits at most three attempts and appends only the evaluator feedback described in Appendix~\ref{app:prompts}. Reference solutions, manufactured target solutions, calibration baselines, hidden solver settings, and case-specific pass/fail thresholds are never included in the agent prompt.

\paragraph{Execution environment.}
Generated solvers are executed in the benchmark sandbox with pinned FEM-library dependencies and the same resource limits used during calibration. Runtime measurements are wall-clock times measured inside this execution environment. Model inference is performed outside the sandbox and is accounted for separately in Appendix~\ref{app:compute_budget}. The experiments reported in this paper were run on a single-node CPU server with 32 visible CPU cores and 62~GB memory. Table~\ref{tab:env_details} summarizes the hardware and software configuration used in the reported experiments.

\begin{table}[h]
\centering
\caption{Hardware and software environment for the reported experiments.}
\label{tab:env_details}
\small
\begin{tabular}{ll}
\toprule
\textbf{Component} & \textbf{Details} \\
\midrule
\multicolumn{2}{l}{\textit{Hardware}} \\
CPU model    & AMD EPYC 7K62 48-Core Processor \\
Visible CPUs & 32 \\
Memory       & 62~GB \\
\midrule
\multicolumn{2}{l}{\textit{Base system}} \\
OS           & Ubuntu 22.04 \\
Python       & 3.11 \\
GCC          & 11.4 \\
MPICH        & conda-forge build \\
PETSc / petsc4py & 3.24.x \\
\midrule
\multicolumn{2}{l}{\textit{DOLFINx track}} \\
DOLFINx      & 0.10.0 \\
pygmsh / meshio & 7.1.x / 5.3.x \\
\midrule
\multicolumn{2}{l}{\textit{Firedrake track}} \\
Container image & \texttt{firedrakeproject/firedrake:latest} evaluated during the 2025 release cycle \\
OS           & Ubuntu 22.04 base image \\
Python       & Bundled with the Firedrake image \\
\midrule
\multicolumn{2}{l}{\textit{deal.II track}} \\
Container image & \texttt{dealii/dealii:v9.7.1-jammy} \\
deal.II      & 9.7.1 \\
OS           & Ubuntu 22.04 base image \\
Python       & System Python 3.x \\
\midrule
\multicolumn{2}{l}{\textit{Execution limits}} \\
Per-case timeout & 300~s wall-clock \\
Internet access  & Disabled inside execution sandbox \\
Filesystem       & Isolated per-case working directory \\
\bottomrule
\end{tabular}
\end{table}

\paragraph{Logged artifacts.}
For each evaluated submission, the benchmark stores the prompt, raw model response, extracted source code, execution stdout and stderr, produced \texttt{solution.npz} or equivalent binary output, \texttt{meta.json}, measured runtime, computed error, and staged verdict. For failed submissions, the same logging convention is used whenever an artifact is available. This makes execution failures, artifact-format failures, accuracy failures, and runtime failures auditable after the run.

\paragraph{Replication protocol.}
To reproduce a reported table or figure, a third party should run the same benchmark version, model set, FEM-library tracks, prompt tier, and evaluation setting through \texttt{scripts/run\_benchmark.py}, then aggregate the emitted verdict logs using the released metric scripts. The released case records materialize the prescribed evaluation grids, supported-library declarations, and scoring thresholds, so reproductions of the reported benchmark results do not require regenerating reference or calibration artifacts. Reference generation and calibration scripts are included for auditing and for extending the benchmark with new cases or new FEM-library tracks.

\paragraph{Version control and determinism.}
The benchmark release includes the case JSONL files, schema, prompt templates, evaluator scripts, container recipes, API-access configuration templates, and aggregation scripts used to produce the reported results. Because hosted LLM APIs can change over time, exact replication also depends on provider-side model snapshots and endpoint behavior. We therefore record the API model identifier, provider, request timestamp, decoding configuration, prompt hash, and response hash for each call. For local execution, deterministic behavior depends on library versions, linear solver backends, mesh-generation routines, and hardware-level numerical differences; the containerized tracks and pinned dependency versions are provided to minimize these sources of variation.

\paragraph{Scope of reproducibility claims.}
The reported pass rates are reproducible from the logged artifacts and evaluation scripts. Re-running hosted model inference may produce small differences if providers update model snapshots or decoding implementations, even under temperature-zero decoding. For this reason, the released logs serve as the auditable record for the reported paper results, while the benchmark harness provides the protocol for future model evaluations under the same case and scoring definitions.

\end{document}